\documentclass[sn-mathphys,Numbered]{sn-jnl}% Math and Physical Sciences Reference Style
\usepackage{graphicx}%
\usepackage{multirow}%
\usepackage{amsmath,amssymb,amsfonts}%
\usepackage{amsthm}%
\usepackage{mathrsfs}%
\usepackage[title]{appendix}%
\usepackage{xcolor}%
\usepackage{textcomp}%
\usepackage{manyfoot}%
\usepackage{booktabs}%
\usepackage{algorithm}%
\usepackage{algorithmicx}%
\usepackage{algpseudocode}%
\usepackage{listings}%
\usepackage{geometry}
\usepackage{soul}
\usepackage{adjustbox}
\usepackage{subcaption}
\usepackage{hyperref}
\usepackage{url}
\usepackage{graphicx,wrapfig,lipsum}

\begin{document}

\title[When do Convolutional Neural Networks Stop Learning?]{When do Convolutional Neural Networks Stop Learning?}


\author*[1]{\fnm{Sahan} \sur{Ahmad}}\email{sahmad@tamusa.edu}

\author[2]{\fnm{Gabriel} \sur{Trahan}}\email{gabriel.trahan1@louisiana.edu}
\author[2]{\fnm{Aminul} \sur{Islam}}\email{aminul@louisiana.edu}

\affil*[1]{\orgdiv{College of Business}, \orgname{Texas A\&M University, San Antonio}, \orgaddress{\street{One University Way}, \city{San Antonio}, \postcode{78224}, \state{Texas}, \country{USA}}}

\affil[2]{\orgdiv{Computer Science}, \orgname{Organization}, \orgaddress{\street{University of Louisiana at Lafayette}, \city{Lafayette}, \postcode{70504}, \state{Louisiana}, \country{USA}}}


\abstract{Convolutional Neural Networks (CNNs) have demonstrated outstanding performance in computer vision tasks such as image classification, detection, segmentation, and medical image analysis. In general, an arbitrary number of epochs is used to train such neural networks. In a single epoch, the entire training data---divided by batch size---are fed to the network. In practice, validation error with training loss is used to estimate the neural network's generalization, which indicates the optimal learning capacity of the network. Current practice is to stop training when the training loss decreases and the gap between training and validation error increases (i.e., the generalization gap) to avoid overfitting. However, this is a trial-and-error-based approach which raises a critical question: Is it possible to estimate when neural networks stop learning based on training data? This research work introduces a hypothesis that analyzes the data variation across all the layers of a CNN variant to anticipate its near-optimal learning capacity. In the training phase, we use our hypothesis to anticipate the near-optimal learning capacity of a CNN variant without using any validation data. Our hypothesis can be deployed as a plug-and-play to any existing CNN variant without introducing additional trainable parameters to the network. We test our hypothesis on six different CNN variants and three different general image datasets (CIFAR10, CIFAR100, and SVHN). The result based on these CNN variants and datasets shows that our hypothesis saves 58.49\% of computational time (on average) in training. We further conduct our hypothesis on ten medical image datasets and compared with the MedMNIST-V2 benchmark. Based on our experimental result, we save $\approx$ 44.1\% of computational time without losing accuracy against the MedMNIST-V2 benchmark. Our code is available at https://github.com/PaperUnderReviewDeepLearning/Optimization}




\maketitle

\section{Introduction}\label{sec1}

``Wider and deeper are better'' has become the rule of thumb to design deep neural network architecture~\cite{ExpandNets,Wider2,ResNet18,Wider4,VGG16}. Deep neural networks behave like a ``double-descent'' curve while traditional machine learning models resemble the ``bell-shaped'' curve as deep neural networks have larger model complexity~\cite{doubledescentCurve}. Deep neural networks require a large amount of data to be trained. The data interpolation is reduced in the deep neural network as the data are fed into the deeper layers of the network. However, a  core question remains: Can we predict whether the deep neural network keeps learning or not based on the training data behavior?

%However, how much training data are needed in a deep neural network to gain optimum performance is not well established.

%%Convolutional Neural Network (CNN) gains impressive performance on computer vision tasks~\citep{ResNet18} that has led to their widespread use. Deeper layer-based CNN tends to achieve higher accuracy on vision tasks such as image classification~\citep{CBS}, image segmentation [???citation], object detection[??citation]. In terms of computational time saving, light-weighted CNN architectures are introduced, with a trade-off between speed and accuracy. However, when a CNN architecture reaches its' model capacity and stops significant learning from the training data remains unclear. 

Convolutional Neural Network (CNN) gains impressive performance on computer vision tasks~\citep{CBS}. Specifically, deeper layer-based CNN tends to achieve higher accuracy on vision tasks such as image classification~\citep{CBS}, image segmentation ~\citep{segmentation}, object detection~\citep{YOLO}. Light-weighted CNN variants are introduced for computational time saving, with a trade-off between speed and accuracy. Considering medical image classification, the tasks can vary from binary to multi-class classification. Unlike general image classification, medical image classification gets its images source from medical professionals such as X-rays and MRIs~\cite{medicalClassificationOne}. The medical dataset scales could range from 100 to 150000, and the data modalities can be designed for a specific purpose by maintaining different imaging protocols. Medical image data can be imbalanced. As an example, a specific diagnosis may contain very few positive samples and a large number of negative samples and can produce a model that is biased~\cite{medicalImageDifficultyOne}. Shallow deep models tend not to perform well for medical image classification because the extracted features are often referred to as low-level features; these features lack representation ability for high-level domain concepts, and their generalization ability is poor~\cite{medicalImageDifficultyseven,medicalImageDifficultyEight}. Because of the features needed to be extracted from medical images to obtain precise information about the image, the computational complexity of using a deep neural model is resource expensive~\cite{medicalImageDifficultyThree}. Transfer learning \cite{transformer} and pre-train weights are common in general image based tasks.  However, the substantial differences
between natural and medical images may advise against such knowledge transfer~\cite{medicalClassificationOne, medicalImageDifficultyFive}. Training with a smaller medical image dataset compared to a general image can quickly lead to overfitting. As a result, the medical image dataset needs to be trained for an optimum time to avoid overfitting. However, it remains unclear when a CNN variant reaches its near-optimal learning capacity and stops significant learning from training data.

In general, all training data are fed into a deep neural model as an epoch in the training phase. The current practice uses many epochs (e.g., 200$\sim$500) to train a deep neural model. The selection of optimal epoch numbers to train a deep neural model is not well established. Following are some of the recent works that use different epoch numbers for their experiments:~\cite{transformer} use 186 epochs to  accelerate
the training of transformer-based language models,~\cite{avid} use 256 epochs for their public video dataset to action recognition ,~\cite{pyglove} use 360 epochs for programming AutoML based on symbolic programming,~\cite{bookprediction} use 150 epochs for pretrained sentence embeddings along with various readability scores for book success prediction. Another trend is to pick the same epoch number for a specific dataset or deep neural model. For example, ~\cite{residual} and~\cite{biological} use 200 epochs for CIFAR10 and CIFAR100 datasets. ~\cite{CBS} and~\cite{neuronmerging} use 200 epochs for ResNet and VGG architecture.~\cite{deepwiener} and~\cite{adam} also use 200 epochs for their two simple global hyperparameters that efficiently trade off between latency and accuracy experiment.~\cite{betterset} use 50$\sim$500 epochs as a range for their synthetic image experiments.~\cite{auctionnetwork} use 1000 epochs for their custom dataset. Recently, MedMNIST-V2~\cite{medMNIST2021} introduced a large-scale MNIST-like collection of standardized medical datasets and performed image classification tasks. In the experiment, they use 100 epochs to train ResNet18 architecture for image classification tasks, regardless of the datasets. In short, most deep neural models adapt a safe epoch for their training. 
\begin{figure*}[h]
 \begin{center} 
     \includegraphics[width=\linewidth]{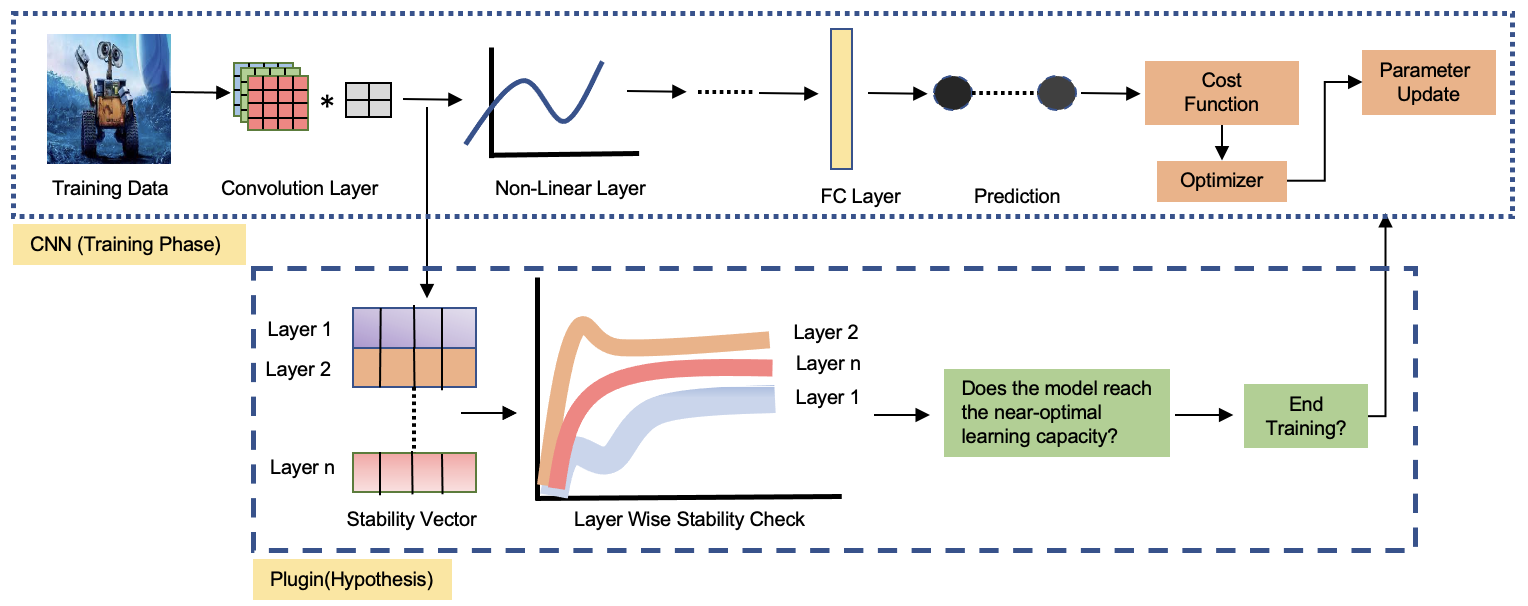}
 \end{center} 
\caption{Top dotted box represents traditional steps of training a CNN variant. At each epoch, our plugin (bottom dotted box) measures data variation after convolution operations. Based on all the layers data variation, the plugin decides the continuity of training.} 
\label{fig:plugIn}
\end{figure*}

Validation data are used alongside training data to estimate the generalization error during or after training~\cite{goodfellow2016deep}. Traditionally, training of the model is stopped when the validation error or generalization gap starts to increase~\citep{goodfellow2016deep}. The generalization gap indicates the model's capability to predict unseen data. However, the current approach of early stopping is based on trial-and-error. This is because it monitors the average loss function on the validation set and continues training until it falls below the value of the training set objective, at which the early stopping procedure is halted~\citep{goodfellow2016deep}. This strategy avoids the high cost of retraining the model from scratch, but it is not as well behaved~\citep{goodfellow2016deep}. For example, the objective on the validation set may never reach the target value, so this strategy is not even guaranteed to terminate~\citep{goodfellow2016deep}. Our research objective is to replace this trial-and-error-based approach with an algorithmic approach to anticipate the near-optimal learning capacity while training a deep learning model. To narrow down the scope of this work, we choose CNN as a member of broader deep learning models.

 %We hypothesize that a layer after convolution reaches its near-optimal learning capacity if the produced data have significantly less variation. We use this hypothesis to identify the epoch where all the layers reach their near-optimal learning capacity, representing the model's near-optimal learning capacity. Thus at each epoch, the proposed hypothesis verifies if the CNN model reaches its near-optimal learning capacity without using validation data. 
 
%\hl{====Modified====}
%After convolution, we hypothesize that a layer reaches its near-optimal learning capacity if the produced data have significantly less variation. Then, we use our hypothesis to identify the epoch where all the layers reach their near-optimal learning capacity, representing the model's near-optimal learning capacity. Thus, our hypothesis predicts the near-optimal epoch number of training a CNN model without using any validation data set. 

Generally, a CNN has some basic functions to conduct the training phase, as illustrated in the top dotted box in Figure~\ref{fig:plugIn}. In practice, a dataset is divided into three parts: Training, validation, and testing. A CNN variant can have convolution, non-linear, and fully connected (FC) layers, and the order of these layers can vary based on the variant. The cost function is the technique of evaluating the performance of a CNN variant, and the optimizer modifies the attributes such as weights and learning rate of a CNN variant to reduce the overall loss and improve accuracy.

We hypothesize (illustrated by bottom dotted box in Figure~\ref{fig:plugIn}) that a layer after convolution operation reaches its near-optimal learning capacity if the produced data have significantly less variation. We use this hypothesis to identify the epoch where all the layers reach their near-optimal learning capacity, representing the model's near-optimal learning capacity. Thus, at each epoch, the proposed hypothesis verifies if the CNN variant reaches its near-optimal learning capacity without using validation data. Our hypothesis terminates the CNN's training when it reaches its near-optimal learning capacity. The hypothesis does not change the learning dynamics of the existing CNN variants or the design of a CNN architecture, cost function, or optimizer. As a result, the hypothesis can be applied to any CNN variant as a plug-and-play after the convolution operation. In summary, any CNN variant that uses training data and/or validation data by multiple epochs can utilize our hypothesis. 

It is worth mentioning that our hypothesis does not introduce any trainable parameter to the network. As a result, our hypothesis can be deployed on any wide and deep or compact CNN variant. The main contributions of this paper can be summarized as:

\begin{itemize}
    \item We introduce a hypothesis regarding near optimal learning capacity of a CNN variant without using any validation data.
    \item We examine the data variation across all the layers of a CNN variant and correlate it to the model's near-optimal learning capacity.
    \item The implementation of the proposed hypothesis can be embodied as a plug-and-play to any CNN variant. 
    \item The proposed hypothesis does not introduce any additional trainable parameter to the network.
    \item To test our hypothesis, we conduct image classification experiments on six CNN variants and three datasets. Utilizing the hypothesis to train the existing CNN variants saves 32\% to 79\% of the computational time.
    \item Finally, we provide a detailed analysis of how the proposed hypothesis verifies the CNN variants' optimal learning capacity.
\end{itemize}

\section{Related Work}
Modern neural networks have more complexity than classical machine learning methods. In terms of bias-variance trade-off for generalization of neural networks, traditional machine learning methods resemble a \textbf{bell shape}, and modern neural networks resemble a \textbf{double descent curve}~\citep{doubledescentCurve}.

In deep neural networks, validation data are used alongside training data to identify the generalization gap~\citep{goodfellow2016deep}. Generalization refers to the model's capability to predict unseen data. The increasing generalization gap indicates that the model is going to overfit. It is recommended to stop training the model at that point. However, this is a trial and error-based approach widely used in the current training strategy. In order to use this strategy, a validation dataset is required.

~\citep{early2016}, ~\citep{early2017},~\citep{early2021} proposed an early stopping method without a validation dataset. However, ~\citep{early2016} and ~\citep{early2017} rely on gradient-related statistics and fail to generalize to more advanced optimizers such as those based on momentum. Both of the works require hyperparameter tuning as well.~\citep{early2021} designed an early stopping method for a specific framework, not a generalized solution.

%\hl{To the best of our knowledge, no previous work investigates at what optimal epoch a CNN model stops learning.}  

There are some CNN architectures that aim to obtain the best possible accuracy under a limited computational budget based on different hardware and/or applications. This results in a a series of works towards light-weight CNN architectures and have speed-accuracy trade-off, including Xception~\citep{xception}, MobileNet~\citep{MobileNet}, ShuffleNet~\citep{shufflenet}, and CondenseNet~\citep{condensenet}. These works use FLOP as an indirect metric to compare computational complexity. ShuffleNetV2~\citep{shufflenet2} uses speed as a direct metric while considering memory access cost and platform characteristics. However, we consider epoch number as a metric to analyze the computational time of training a CNN variant.
%or at what optimal epoch a CNN model reaches its near-optimal learning capacity is not well researched.

The usual practice is to adopt a safe epoch number for a specific dataset and a CNN variant. However, the epoch number selection is random, and an arbitrary safe number is picked for most of the experiments. This inspires us to investigate when a CNN variant almost stops learning significantly from the training data.

\section{Training Behavior of Convolutional Neural Network}

\subsection{Convolutional neural network (CNN)}

To denote the convolutional operation of some kernel $\theta_k$ on some input $X$, we use $\theta_k \circledast X$. In deep learning, a typical CNN is composed of stacked trainable convolutional layers ~\citep{CNN}, pooling layers~\citep{pool}, and non-linearities~\citep{nonlinear}.

In a single epoch $(e)$, the entire training data (i.e., the number of training samples) is sent by multiple iterations $(t)$ with batch size $(N)$. Thus the number of training samples ($D_{\text{train}}$) sent in a single epoch is expressed by the following equation:
\begin{equation}
\label{eq:data}
    D_{\text{train}} = N  \times t
\end{equation}
The input tensor $X$ is organized by batch size $N$, channel number $c$, height $h$, and width $w$ as $X(N,c,h,w)$. A typical CNN  convolution operation at $n$-th layer and at $t$-th iteration can be mathematically represented by Equation~\ref{eq:conv}, where $\theta_k$ are the learned weights of the kernel.

\begin{equation} 
    \label{eq:conv}
     X_n^t =(\theta_k \circledast X_{n-1}^t)
\end{equation}
\subsection{Stability vector}
\label{sec:stabilityVector}
\begin{figure}
\centering 
\includegraphics[scale=0.45]{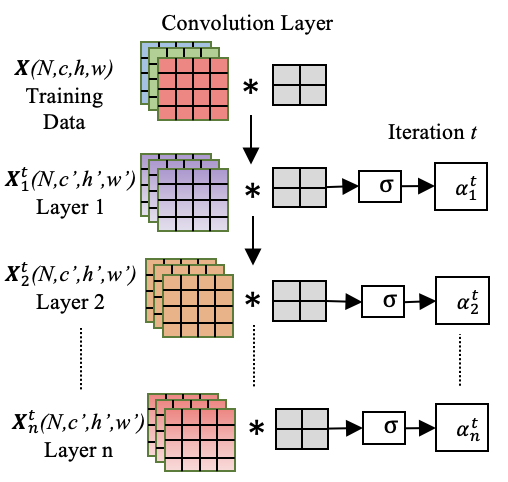}
\caption{At $t$-th iteration, the process of computing stability values $\alpha_1^t, \alpha_2^t, \ldots,\alpha_n^t$ for 1 to $n$ layers.}
\label{fig:stablescalar}
\end{figure}
In the training phase, we examine whether the CNN model keeps learning or not by measuring data variation after convolution operation. To do that we introduce the concept of stability value and stability vector. After the convolution operation at $t$-th iteration and $n$-th layer, we measure the stability value (element of a stability vector) $\alpha_n^t$ by computing the standard deviation value of \ $X_n^t$ as $\alpha_n^t = \sigma (X_n^t)$. The process of constructing stability values is shown in Figure~\ref{fig:stablescalar}.  

\begin{figure}{r}
\centering
    \includegraphics[scale=0.4]{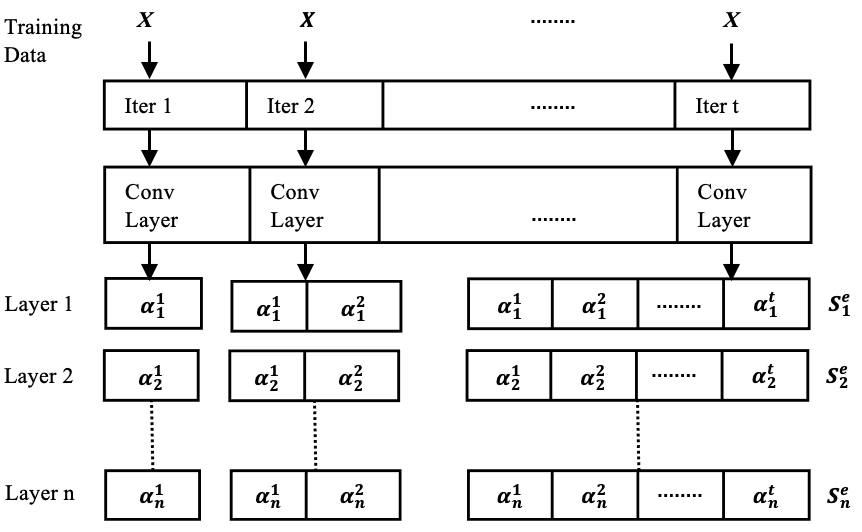}

\caption{At $e$-th epoch, the process of constructing stability vectors $S_1^e$, $S_2^e$, \ldots, $S_n^e$ for 1 to $n$ layers.}
\label{fig:stablesvector}
\end{figure}

At $e$-th epoch and at $n$-th layer, we construct stability vector $S_n^e = [\alpha_n^1, \alpha_n^2, \ldots, \alpha_n^t]$ by computing the stability values for all the iterations $(t)$. At $e$-th epoch, the process of constructing stability vectors for all the layers (i.e., layers 1 to $n$) after $t$ number of iterations is shown in Figure~\ref{fig:stablesvector}. Thus at each epoch, we have $n$ number of stability vectors (i.e., based on the number of layers) with size $t$ (i.e., the number of iterations).

\subsection{Layer and model stability}

Significantly less data variation of a particular layer's stability vectors for consecutive epochs indicates that that layer of the CNN become stable (i.e., fails to learn significant information from training data). When all the layers of the model get stable, it implies the possibility that the model reaches its near-optimal learning capacity.

In order to measure the data variation of a layer for two consecutive epochs, at first we compute the mean of stability vector, $\mu_{n}^e$, at $e$-th epoch and $n$-th layer by the following equation:

%After $t$ iterations at $e$-th epoch and $n$-th layer, we have the stability vector $S_n^e = [\alpha_{n1}, \alpha_{n2}, \ldots, \alpha_{nt}]$.

\begin{equation}
    \label{eq:mu}
    \mu_{n}^e = \frac{1}{t} \sum_{i=1}^{t}{\alpha_n^i}
\end{equation}

 We define a function $p^r$ that rounds a number to $r$ decimal places. For example, if $\mu_{n}^e=1.23456$, $p^2(\mu_{n}^e)$ will return 1.23.
 At $n$-th layer, we compare the mean \st{of} stability vector of epoch $e$ with its previous one by rounding to decimal places $r$ by using the following equation:
\begin{equation}
    \label{eq:precision}
    \delta_n^e =p^r(\mu_{n}^e) - p^r(\mu_{n}^{e-1})
\end{equation}
 At $e$-th epoch and $n$-th layer, if $\delta_n^e$ equals zero, we consider that the $n$-th layer is stable at $e$-th epoch. If all the layers show the stability by using $\sum_{i=1}^n \delta_i^e =0$ indicates the possibility that the CNN model gets stable (i.e., reaches its near-optimal learning capacity) at $e$-th epoch.
It also means that the model does not extract any more significant information from the training data. To make sure that the CNN model reaches its near-optimal learning capacity, we repeat using $\sum_{i=1}^n \delta_i^e =0$ for two more epochs (i.e., epochs $e+1$, and $e+2$). If the result remains the same, we conclude that the model reaches its near-optimal learning capacity and we terminate the training phase. The trained model is now ready for the testing environment.

All \st{the} variables we use in our hypothesis are not trained via back-propagation and do not introduce any trainable parameter to the network.

\subsubsection{A Walk-through Example of Model Stability on ResNet18 Architecture (using CIFAR100 dataset)} 

\begin{table}
\caption{$p^2(\mu_{n}^e)$ values across epoch 73 to 76 for ResNet18 on CIFAR100 dataset ($p^2(\mu_{n}^e)$ values are from Figure~\ref{fig:stable}) }
\label{resnet18table}
%\begin{center}
\centering
\renewcommand{\tabcolsep}{1pt}
\fontsize{8.4}{11}\selectfont
\begin{tabular}{lccccccc}
\hline

\multicolumn{1}{c}{ Layer} 
&\multicolumn{1}{c}{ $p^2(\mu_{n}^{73})$}
&\multicolumn{1}{c}{ $p^2(\mu_{n}^{74})$}
&\multicolumn{1}{c}{ $\delta_n^{74}$}
&\multicolumn{1}{c}{ $p^2(\mu_{n}^{75})$} 
&\multicolumn{1}{c}{ $\delta_n^{75}$}
&\multicolumn{1}{c}{ $p^2(\mu_{n}^{76})$}
&\multicolumn{1}{c}{ $\delta_n^{76}$}
\\ \hline
\hspace{5pt}1         &0.14  &0.14 &0 & 0.14 &0 &0.14  &0\\
\hspace{5pt}5         &0.19  &0.19 &0 & 0.19 &0 &0.19  &0\\
\hspace{5pt}9         &0.14  &0.14 &0 & 0.14 &0 &0.14  &0\\
\hspace{5pt}13        &0.09  &0.09 &0 & 0.09 &0 &0.09  &0\\
\hspace{5pt}18        &0.44  &0.44 &0 & 0.44 &0 &0.44  &0\\
\hline
\end{tabular}
%\end{center}
\end{table}
In CIFAR100 dataset, the total number of training sample is 50000. We consider 64 as the batch size for training (i.e., $N=64$). So, in each epoch $(e)$, the iteration number is $\frac{50000}{64} = 782$ (i.e., $t=782$).

At $e$-th epoch and $n$-th layer, the first iteration constructs the first element (i.e., $\alpha_n^1$) of stable vector $S_n^e$. In ResNet18 architecture, at epoch $e$, there are 18 layers and for each layer we construct one stability vector, so we have in total 18 stability vectors (i.e., $S_1^e, S_2^e, \ldots, S_{18}^e$). The length of each stability vector is 782 because each epoch consists of 782 iterations (Figure~\ref{fig:stablesvector}). Table~\ref{resnet18table} shows the $p^2(\mu_{n}^e)$ values for epoch 74 to 76. As the $\delta_n^e$ is 0 for three consecutive epochs, our hypothesis terminates the ResNet18 training on CIFAR100 dataset at epoch 76.

\section{Experiments}

In this section, we empirically evaluate the effectiveness of our hypothesis on six different CNN variants such as ResNet18~\citep{ResNet18}, ResNet18+CBS~\citep{CBS}, CNN~\citep{CNN}, CNN+CBS~\citep{CBS}, VGG16~\citep{VGG16}, and VGG16+CBS~\citep{CBS}. We test these CNN variants on three different datasets (i.e., CIFAR10, CIFAR100 ~\citep{CIFAR10}, and SVHN~\citep{SVHN}). MedMNIST-V2~\cite{medMNIST2021} introduce a large scale MNIST-like collection of standardized medical datasets. MedMNIST-V2 use ResNet18 \cite{ResNet18} architecture on the standardized medical dataset for image classification task. We evaluate the effectiveness of our hypothesis against the benchmark of MedMNIST-V2 and also add CNN~\cite{CNN} architecture. Finally we analyze the computational time saving (CTS) and Top-1 classification accuracy by utilizing our hypothesis. We further provide an ablation study to analyze the influence of our strategy. 

\subsection{CNN Variants, datasets, and tasks}
\subsubsection{General Image}

To evaluate our hypothesis, we perform the image classification task on two standard vision datasets, CIFAR10 and CIFAR100, containing images for 10 and 100 classes, respectively. SVHN, the other dataset, is a digit recognition dataset that consists of natural images of the 10 digits collected from the street view. Table~\ref{datasetTable} shows more details about these datasets.

CNN demonstrated remarkable performance in computer vision tasks. Both ResNet~\citep{ResNet18} and VGG~\citep{VGG16} are based on CNN architecture and have different variations based on the number of layers. We consider ResNet18~\citep{ResNet18} and VGG16~\citep{VGG16} variations in our experiment. Curriculum by Smoothing~\citep{CBS}(CBS) is a general method for training CNNs, which can be applied to any CNN variant. CBS controls the amount of high-frequency information during the training phase. It augments the training scheme and increases the amount of information in the feature maps so that the network can progressively learn a better representation of the data. CBS applied to CNN variants, such as ResNet18+CBS and VGG16+CBS, improve the accuracy of image classification tasks. We also utilize our hypothesis with ResNet18+CBS and VGG16+CBS variants in the experiment.
\begin{table}
\caption{General Image Dataset}
\label{datasetTable}
\centering
\renewcommand{\tabcolsep}{2pt}
\fontsize{8.4}{11}\selectfont
%\begin{adjustbox}{width=.8\textwidth}
\begin{tabular}{lccccc}
\hline 
\multicolumn{1}{c}{ Dataset} 
&\multicolumn{1}{c}{ Batch }
&\multicolumn{1}{c}{ Train.}
&\multicolumn{1}{c}{ Train. } 
&\multicolumn{1}{c}{ Valid. }
&\multicolumn{1}{c}{ Valid. } \\
& Size & Data  & Iter.  & Data & Iter. \\
&  ($N$) & ($D_{\text{train}}$) & ($t_\text{train}$) &($D_{\text{val}}$) &  ($t_\text{val}$)\\
\hline
CIFAR10      &64  &50000 &782  &10000 &157\\
CIFAR100     &64  &50000 &782  &10000 &157\\
SVHN         &64  &73257 &1145 &26032 &407\\

\end{tabular}
%\end{adjustbox}

\end{table}

\subsubsection{Medical Image}
\begin{table*}[htp]
\caption{Medical Image Datasets}
\label{datasetTableM}
\centering
\renewcommand{\tabcolsep}{2.1pt}

%\begin{adjustbox}{width=.8\textwidth}
\begin{tabular}{lcrrrrrr}
\hline 
\multicolumn{1}{c}{ Dataset} 
&\multicolumn{1}{c}{ Batch }
&\multicolumn{1}{c}{ Train.}
&\multicolumn{1}{c}{ Train. } 
&\multicolumn{1}{c}{ Valid. }
&\multicolumn{1}{c}{ Valid. } 
&\multicolumn{1}{c}{ Test. }
&\multicolumn{1}{c}{ Test. } \\
& Size & Data  & Iter.  & Data & Iter. & Data & Iter.\\
&  ($N$) & ($t_{data}$) & ($t_i$) &($v_{data}$) &  ($v_i$)&($f_{data}$) &  ($f_i$)\\
\hline
Pneumonia Detection                     &64  &24000 & 375  &6000 &94 & 4527 &  71\\
Retinal OCT (OCTMNIST)                  &64  &97477 & 1524 &10832 & 170 & 1000 & 17\\
Blood Cell Microscope (BloodMNIST)       &64  &11959 & 187 &1712 & 27 & 3421 & 54\\
Kidney Cortex Microscope (TissueMNIST)  &64  &165466 & 2586 &23640 & 370 & 47280 & 739\\
Abdominal CT Axial (OrganAMNIST)         &64  &34581 & 541 &6491 & 102 &17778 & 278\\
Abdominal CT Coronal (OrganCMNIST)       &64  &13000 & 204 & 2392 & 38 & 8268 &  130\\
Abdominal CT Sagitta (OrganSMNIST)       &64  &13940 & 218 &2452 & 39 & 8829 & 138\\
Colon Pathology (PathMNIST)             &64  &89996 & 1407 &10004 & 157 & 7180 & 113\\
Dermatoscope (DermaMNIST)               &64 & 7007 & 110 & 1003 & 17 & 2005 & 32 \\
Fundus Camera (RetinaMNIST)             &64 &1080 & 17 & 120 & 2 & 400 & 7 \\
\end{tabular} %\vspace{-10pt}
%\end{adjustbox}
%\end{center}
\end{table*}

To evaluate the effectiveness of our hypothesis, we perform the medical image classification task on the following 10 standard medical vision datasets:  

\textbf{Pneumonia Detection} dataset contains images of frontal view chest radiograph (X-ray). It contains an X-ray of pneumonia or without pneumonia symptoms, making it a binary classification problem.

\textbf{PathMNIST} contains images from hematoxylin and eosin-stained histological images. The dataset is comprised of 9 types of tissues which forms a multi-class classification task.

\textbf{OCTMNIST} contains optical coherence tomography (OCT) images for retinal diseases. This dataset is comprised of four diagnosis categories.

\textbf{BloodMNIST} dataset is based on normal cells, captured from individuals without infection, hematologic or oncologic diseases and free of any pharmacologic treatment at the moment of blood collection. It is organized into eight classes.

\textbf{TissueMNIST} dataset is based on human kidney cortex cells, segmented from three reference tissue specimens and organized into eight categories.

\textbf{Organ\{A,C,S\}MNIST} contains computed tomography (CT) images from Liver Tumor Segmentation benchmarks. It contains eleven annotated body organs renamed as OrganMNIST\{Axial, Coronal, Sagittal\}.

\textbf{DermaMNIST} is a large collection of multi-source dermatoscopic images of common pigmented skill lesions. The images are categorized as seven different diseases.

\textbf{RetinaMNIST} contains retina fundus images. The task is based on five level grading of ordinal regression of diabetic retinopathy severity.

Table~\ref{datasetTableM} shows more details about these datasets. CNN demonstrated remarkable performance in computer vision tasks. ResNet~\cite{ResNet18} is based on CNN architecture and have different variations based on the number of layers. We consider ResNet18~\cite{ResNet18} variation in our experiment as MedMNIST-V2~\cite{medMNIST2021} also uses ResNet18 to create a benchmark. We also use another base CNN variant.

\subsection{Computational time  saving (CTS)}
Let the total number of epochs required to train a CNN variant be $E$. Then, based on Equation~\ref{eq:data}, we compute the total iteration (training iteration and validation iteration) needed in $E$ epochs to train a CNN variant by the following equation\footnote{Symbols are defined in Table~\ref{datasetTable}.}:
\begin{equation}
    \label{eq:iter}
    t_{total} =E (\frac{D_{train}}{N} + \frac{D_{val}}{N})
\end{equation}

Computational time saving (CTS) between model $m_1$ and $m_2$ defines how much less time (i.e, percentage decrease) in terms of total iteration (i.e., $t_{total}$) required by $m_1$ to complete the training than $m_2$. For an example, in order to train ResNet18 architecture on CIFAR100 dataset, the total number of iteration required based on Equation~\ref{eq:iter} is $200((50000 / 64) + (10000 / 64) = 187800$. At 76 epoch, our hypothesis anticipates that ResNet18 reaches its near-optimal learning capacity and terminates the training. By utilizing our hypothesis in ResNet18 on the CIFAR100 dataset requires $76(50000 / 64) =59432$ iterations to train which saves $\frac{(187800-59432)}{187800}$=68.35\% computation and gains $\pm$ 0.30 top-1 classification accuracy.

\subsubsection{CTS of General Image Classification }

We consider 200 epochs as the benchmark epoch number. CBS~\citep{CBS} use 200 epochs in their experiments. \cite{residual} use 200 epochs on CIFAR10 and CIFAR100 datasets.~\cite{neuronmerging} use 200 epochs on VGG and ResNet variants. We use CBS, VGG, and ResNet architectures on CIFAR10, CIFAR100 datasets and compare the CTS based on 200 epochs for all of our experiments (i.e., six CNN architectures and three datasets). We keep the batch size constant (i.e., 64) for all the datasets. That is, in one iteration, the model uses 64 samples.

\begin{table*}[htb]
\caption{Computational time saving (CTS) in percentage and Top-1 classification accuracy (Acc.) on CIFAR10, CIFAR100, SVHN datasets. The \textbf{bold} numbers represent better scores.}
\label{expTable}
%\begin{adjustbox}{width=1.0\textwidth}
\renewcommand{\tabcolsep}{0.2pt}
\fontsize{5.5}{8}\selectfont

\begin{tabular}{l cccccccccccc}
\hline
DataSet &
\multicolumn{4}{|c|}{CIFAR10} &
\multicolumn{4}{c|}{CIFAR100} &
\multicolumn{4}{c}{SVHN} \\
\hline
Model & Train. & Total & CTS & Acc. &
        Train. & Total & CTS & Acc. &
        Train. & Total & CTS & Acc.\\
 & Epoch & Iter. & (in \%) &  &
        Epoch & Iter. & (in \%) &  &
        Epoch & Iter. & (in \%) & \\        
\hline
CNN &200 &187800 & 0 & \textbf{80.4}  $\!\!\pm\!\!$ 0.2            &200 &187800 & 0 &48.2 $\!\!\pm\!\!$ 0.2       &200 &310400 &0 &89.6 $\!\!\pm\!\!$ 0.2  \\
CNN+Our &\textbf{78} & \textbf{60996} & \textbf{67.5} &79.5 $\!\!\pm\!\!$ 0.2       & \textbf{123} & \textbf{96186} & \textbf{48.8} & \textbf{49.2} $\!\!\pm\!\!$ 0.2       & \textbf{99} & \textbf{113355} & \textbf{63.5} & \textbf{89.8} $\!\!\pm\!\!$ 0.2  \\
CNN+CBS &200 &187800 & 0 & \textbf{77.3} $\!\!\pm\!\!$ 0.2        &200 &187800 & 0 &\textbf{46.5} $\!\!\pm\!\!$ 0.2           &200 &310400 &0 &\textbf{89.4} $\!\!\pm\!\!$ 0.2  \\
CNN+CBS+Our &\textbf{128} &\textbf{100096} &\textbf{46.7} &77.2 $\!\!\pm\!\!$ 0.2  &\textbf{139} &\textbf{108698} &\textbf{42.1} &46.4 $\!\!\pm\!\!$ 0.2 &\textbf{134} &\textbf{153430} &\textbf{50.6} &89.2 $\!\!\pm\!\!$ 0.2  \\

ResNet18 &200 &187800 &0 &\textbf{89.3}  $\!\!\pm\!\!$ 0.3            &200 &187800 &0 &64.3 $\!\!\pm\!\!$ 0.3      &200 &310400 &0 &\textbf{95.0} $\!\!\pm\!\!$ 0.2  \\
ResNet18+Our &\textbf{59} &\textbf{46138} &\textbf{75.3} &89.0  $\!\!\pm\!\!$ 0.3       &\textbf{76} &\textbf{59432} &\textbf{68.3}&\textbf{64.6} $\!\!\pm\!\!$ 0.3     &\textbf{56} &\textbf{64120} &\textbf{79.3} &94.3 $\!\!\pm\!\!$ 0.2  \\
ResNet18+CBS &200 &187800 &0 &\textbf{89.3}  $\!\!\pm\!\!$ 0.3        &200 &187800 &0 &\textbf{65.8} $\!\!\pm\!\!$ 0.3      &200&310400 &0 &\textbf{96.1} $\!\!\pm\!\!$ 0.2  \\
ResNet18+CBS+Our &\textbf{65} &\textbf{50830} &\textbf{72.8} &89.0  $\!\!\pm\!\!$ 0.3   &\textbf{74} &\textbf{57868} &\textbf{69.2} &65.3 $\!\!\pm\!\!$ 0.3    &\textbf{63} &\textbf{72135} &\textbf{76.7} &95.9 $\!\!\pm\!\!$ 0.2  \\

VGG16 &200 &187800 &0 &\textbf{82.0}  $\!\!\pm\!\!$ 0.2               &200 &187800 &0 &\textbf{48.8} $\!\!\pm\!\!$ 0.3        &200 &310400 &0 &\textbf{93.8} $\!\!\pm\!\!$ 0.2  \\
VGG16+Our &\textbf{109} &\textbf{85238} &\textbf{54.6} &81.7  $\!\!\pm\!\!$ 0.2         &\textbf{163} &\textbf{127466} &\textbf{32.1} &48.0 $\!\!\pm\!\!$ 0.3    &\textbf{113} &\textbf{129385} &\textbf{58.3} &93.6 $\!\!\pm\!\!$ 0.2 \\
VGG16+CBS &200 &187800 &0 &\textbf{83.6}  $\!\!\pm\!\!$ 0.3              &200 &187800 &0 &49.1 $\!\!\pm\!\!$ 0.3            &200 &310400 &0 &94.2 $\!\!\pm\!\!$ 0.2 \\
VGG16+CBS+Our &\textbf{109} &\textbf{85238} &\textbf{54.6} &83.5  $\!\!\pm\!\!$ 0.3    &\textbf{148} &\textbf{115736} &\textbf{38.4} &\textbf{50.4} $\!\!\pm\!\!$ 0.3    &\textbf{125} &\textbf{143125} &\textbf{53.9} &\textbf{94.5} $\!\!\pm\!\!$ 0.2  \\
\hline
\end{tabular}
%\end{adjustbox}
\end{table*}

\subsection{CTS of Medical Image Dataset}
We consider 100 epochs as the benchmark epoch number (i.e., early stop) as MedMNIST-V2~\cite{medMNIST2021} use 100 epochs in their experiments. We use CNN, and ResNet18 architectures on Pneumonia Detection, OCTMNIST, BloodMNIST, TissueMNIST, OrganAMNIST, OrganCMNIST, OrganSMNIST, PathMNIST, DermaMNIST and RetinaMNIST datasets and compare the CTS based on 100 epochs for all of our experiments (i.e., Two CNN architectures and ten datasets). We keep the batch size constant (i.e., 64) for all the datasets. That is, in one iteration, the model uses 64 samples.

\begin{table*}[htb]
\footnotesize
\caption{Computational time saving (CTS) in percentage and Top-1 classification accuracy (Acc.) for ResNet18 and CNN on  Pneumonia Detection, OCTMNIST, BloodMNIST, TissueMNIST, OrganAMNIST, OrganCMNIST, OrganSMNIST, PathMNIST, DermaMNIST and RetinaMNIST datasets. The \textbf{bold} numbers represent better scores.} 
\label{expTableM}
%\begin{adjustbox}{width=1.0\textwidth}
\renewcommand{\tabcolsep}{2.9pt}

\begin{tabular}{l|ccccc|ccccc}
\hline
  & \multicolumn{10}{|c}{Model} \\
\hline
 & \multicolumn{5}{|c|}{ResNet18 (MedMNIST-V2)} &\multicolumn{5}{c}{ResNet18(MedMNIST-V2) + CL} \\
\hline
Dataset & Train. & Val.&  Total & CTS & Accuracy & Train. & Val. & Total & CTS & Accuracy \\
 & Epoch & Epoch & Iter. & (in \%) &  & Epoch & Epoch & Iter. & (in \%) &  \\
\hline

Pneumonia   & 100 & 100 & 46K & 0 & 76.5$\pm$ 0.2 & 42 & 0 & 15K & \textbf{66.4} & \textbf{76.9$\pm$ 0.2} \\
OCTMNIST    & 100 & 100 & 169K & 0 & 82.2 $\pm$ 0.3 & 61 & 0 & 92K & \textbf{45.1} & \textbf{82.8 $\pm$ 0.3}\\
BloodMNIST  & 100 & 100 & 214K & 0 & \textbf{95.8 $\pm$ 0.3} & 71 & 0 & 132K & \textbf{37.9} & 95.4 $\pm$ 0.3\\
TissueMNIST & 100 & 100 & 295K & 0 & \textbf{67.6 $\pm$ 0.4} & 94 & 0 & 243K & \textbf{17.7} & 67.3 $\pm$ 0.4 \\
OrganAMNIST & 100 & 100 & 64K & 0 & 93.5 $\pm$ 0.2 & 63 & 0 & 34K & \textbf{46.9} & \textbf{94.3 $\pm$ 0.2}\\
OrganCMNIST & 100 & 100 & 24K & 0 & 90.0 $\pm$ 0.2 & 53 & 0 & 10K & \textbf{55.3} & \textbf{91.3 $\pm$ 0.2}\\
OrganSMNIST & 100 & 100 & 25K & 0 & 78.2 $\pm$ 0.2 & 60 & 0 & 13K & \textbf{49.1} & \textbf{79.6 $\pm$ 0.2}\\
PathMNIST   & 100 & 100 & 156K & 0 & \textbf{90.7 $\pm$ 0.3} & 59 & 0 & 83K & \textbf{46.9} & \textbf{90.7 $\pm$ 0.3}\\
DermaMNIST  & 100 & 100 & 12K & 0 & 73.5 $\pm$ 0.2  & 60 & 0 & 6K & \textbf{48.0} & \textbf{74.3 $\pm$ 0.2}\\
RetinaMNIST & 100 & 100 & 1.9K & 0 & \textbf{52.4 $\pm$ 0.2} & 81 & 0 & 1.3K & \textbf{27.5} & 52.2$\pm$ 0.2\\
\hline
 & \multicolumn{5}{|c|}{CNN} & \multicolumn{5}{c}{CNN + CL} \\
\hline

Pneumonia   & 100 & 100 & 46K & 0 & \textbf{69.1 $\pm$ 0.2} & 65 & 0 & 24K & \textbf{48.0}  & 69.0 $\pm$ 0.2\\
OCTMNIST    & 100 & 100 & 169K & 0 & 70.5 $\pm$ 0.3 & 46 & 0 & 70K & \textbf{58.6} & \textbf{70.7 $\pm$ 0.3} \\
BloodMNIST  & 100 & 100 & 214K & 0 & 92.6 $\pm$ 0.3 & 64 & 0 & 119K & \textbf{44.0} & \textbf{92.7 $\pm$ 0.3}\\
TissueMNIST & 100 & 100 & 295K & 0 & \textbf{58.6 $\pm$ 0.4} & 101 & 0 & 261K & \textbf{11.6} & \textbf{58.6 $\pm$ 0.4}\\
OrganAMNIST & 100 & 100 & 64K & 0 & 91.4 $\pm$ 0.2 & 92 & 0 & 49K & \textbf{22.5} & \textbf{91.6 $\pm$ 0.2}\\
OrganCMNIST & 100 & 100 & 24K & 0 & \textbf{90.3 $\pm$ 0.2} & 63 & 0 & 12K & \textbf{46.8} & 90.2 $\pm$ 0.2\\
OrganSMNIST & 100 & 100 & 25K & 0  & 78.0 $\pm$ 0.2 & 65 & 0 & 14K& \textbf{46.8} & \textbf{78.1 $\pm$ 0.2}\\
PathMNIST   & 100 & 100 & 156K & 0 & \textbf{83.2 $\pm$ 0.3} & 73 & 0 & 10K & \textbf{34.3}  &83.1 $\pm$ 0.3\\
DermaMNIST  & 100 & 100 & 12K  & 0 & 74.0 $\pm$ 0.2 & 66 &0 & 7K & \textbf{42.8} & \textbf{ 74.2 $\pm$ 0.2}\\
RetinaMNIST & 100 & 100 & 1.9K  & 0 & 50.8 $\pm$ 0.2 & 98 & 0 & 1.6 K & \textbf{12.3} & \textbf{51.0 $\pm$ 0.2}\\
\hline
\end{tabular}
%\end{adjustbox}
\end{table*}

\subsection{Ablation study}

The ablation study results are summarized in Table~\ref{expTable} and Table~\ref{expTableM}. To evaluate the Computational time saving (CTS) and Top-1 classification accuracy (Acc.) for general image classification, we run 36 experiments, 18 of which are conducted without our hypothesis, and 18 of which are conducted with our hypothesis. 200 epoch number is considered safe by the respective researchers on these three datasets and across the six CNN variants. For all 18 experiments, our hypothesis anticipates the near-optimal learning capacity of CNN variants which require significantly less than 200 epochs to train.   

By using our hypothesis, computational time saving ranges from 32.12\% to 79.34\%. On average, we save 58.49\% computational time based on the 18 experiments. We report the mean accuracy over five different seeds. All experimental results for CIFAR10, CIFAR100, and SVHN are listed in Table~\ref{expTable} where we report the top-1 classification accuracy.

To evaluate the CTS and Top-1 classification accuracy (Acc.) for medical image classification,  we run 40 experiments, 20 of them are conducted without our curriculum learning, and the rest 20 are with our hypothesis. 100 epoch is considered safe by the MedMNIST-V2 on these ten datasets and across the two CNN variants. For 19 experiments out of 20, our hypothesis estimates the optimal learning capacity of the two CNN variants which require less than 100 epochs to train.

Computational time-saving ranges from 17.7\% to 66.4\% for ResNet18 and 11.6\% to 58.6\% for CNN. On average, we save 44.1\% and 36.6\%  computational time on ResNet18 and CNN, respectively  based on the 20 experiments. We report the mean accuracy over five different seeds. All experimental results for these 10 datasets are listed in Table~\ref{expTableM} where we report the top-1 classification accuracy.

In our experiments, the dataset sizes range from small to large. For example the dataset size of RetinaMNIST, DermaMNIST, and BloodMNIST is small. Pneumonia detection, OrganAMNIST, OrganCMNIST, and OrganMNIST can be considered mid-size datasets. OCTMNIST, TissueMNIST, and PathMNIST contain a large amount of training data. Regardless of the dataset size, our hypothesis estimates the optimal training epoch required for ResNet18 and CNN. For optimization, we use stochastic gradient descent (SGD) with the same learning rate scheduling, momentum, and weight decay as stated in the original papers~\cite{medMNIST2021,ResNet18}, without hyper-parameter tuning. The task objective for all image classification experiments is a standard unweighted multi-class cross-entropy loss~\cite{medMNIST2021}.

%the maximum accuracy we gain for the VGG16+CBS architecture on CIFAR100 dataset is 1.3 and the maximum accuracy drop for CNN on the CIFAR10 dataset  is 0.94. On average, the accuracy drops 0.106, based on the 36 experiments.

\subsection{Generalization and near-optimal learning capacity}
In our experiments, we work with six different CNN variants. For optimization, we use stochastic gradient descent (SGD) with the same learning rate scheduling, momentum and weight decay as stated in the original papers~\citep{CBS}~\citep{ResNet18}~\citep{VGG16}, without hyper-parameter tuning. The task objective for all image classification experiments is a standard unweighted multi-class cross-entropy loss ~\citep{CBS}.

\begin{figure}
\renewcommand{\tabcolsep}{0.1pt}
\begin{tabular}{cc}
 \includegraphics[width=.50\linewidth]{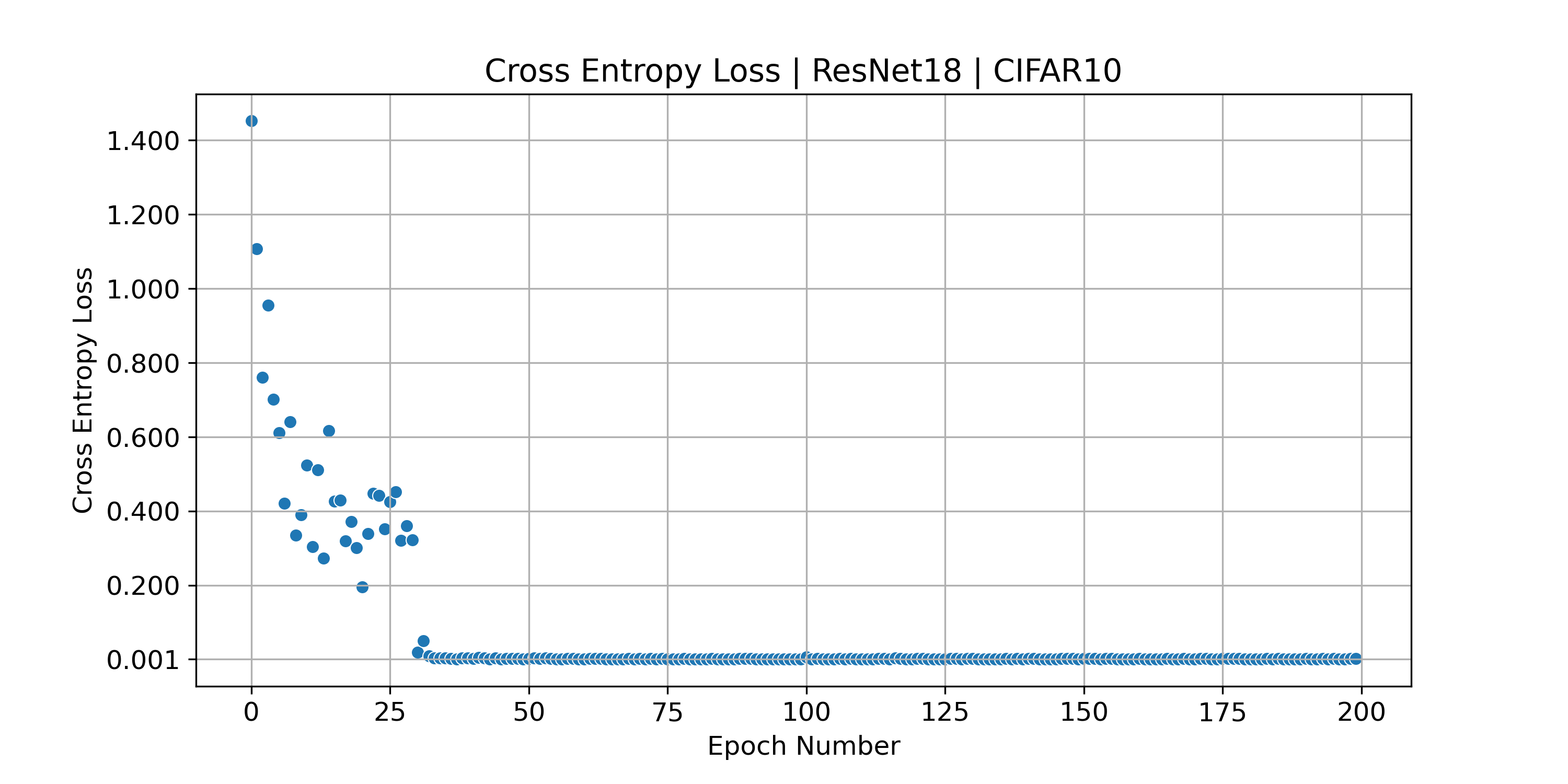} &  
 \includegraphics[width=.50\linewidth]{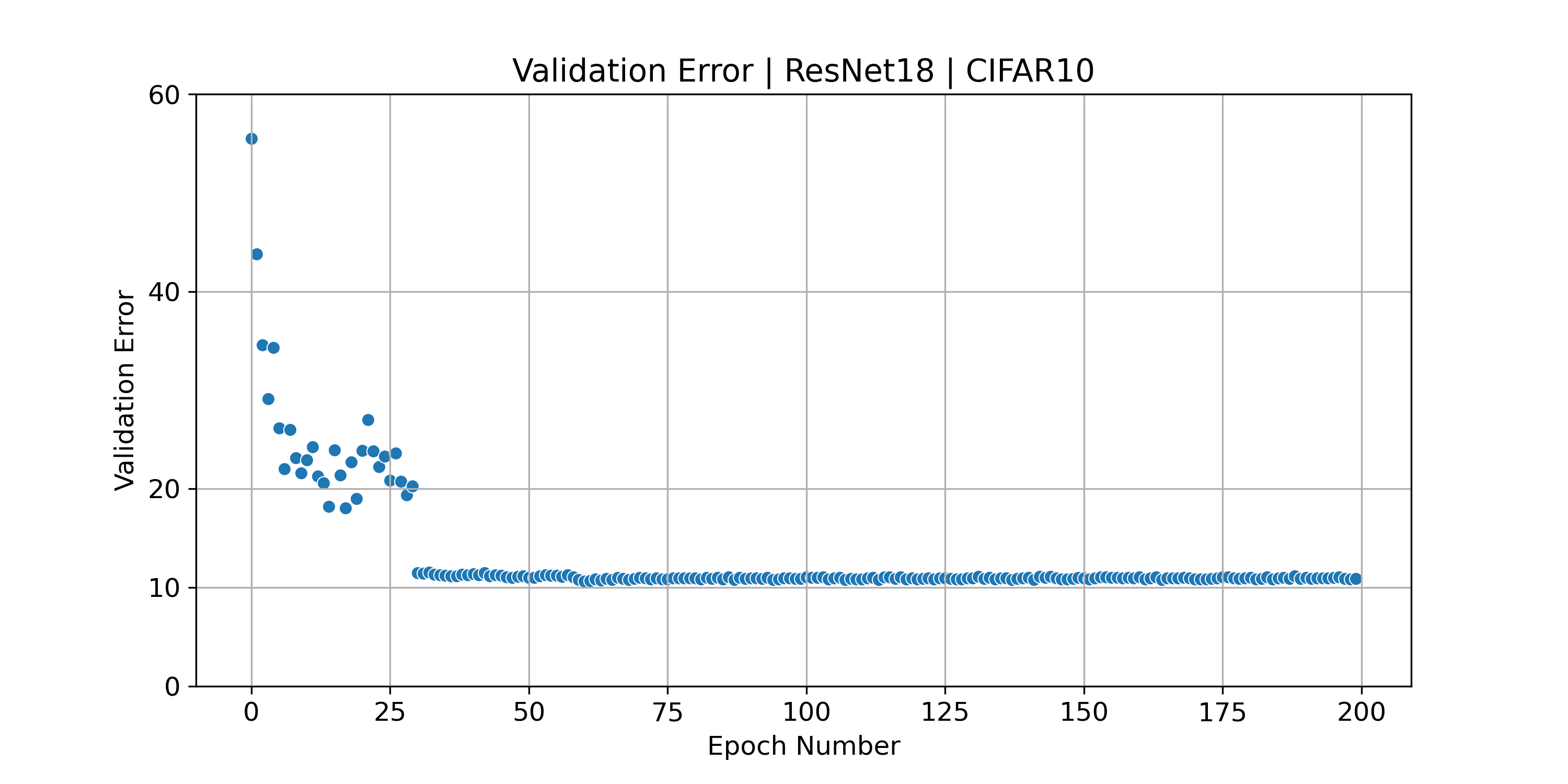} \\
\end{tabular}
\caption{The cross entropy loss (top) and the validation error (bottom) are shown up to 200 epochs for ResNet18 on the CIFAR10 dataset.}
\label{fig:trainValErro}
\end{figure}

~\citep{GenerailzationICLR} conducts an empirical study about generalization by using thousands of models with various fully-connected architectures, optimizers, and other hyper-parameter on image classification datasets. For the image classification task, based on the wide range of experiments on the CIFAR10 dataset, ~\citep{GenerailzationICLR}  concluded that train loss does not correlate well with generalization. In  the 18 experiments without our hypothesis (i.e., using validation data), we observe similar behavior in the training phase. As an example, Figure~\ref{fig:trainValErro} shows the cross-entropy (CE) loss and validation error on the CIFAR10 dataset for ResNet18 architecture. The CE loss and validation error reduce in the beginning phase of training. However, after that, the generalization gap (i.e., the increase of validation error with CE loss) does not significantly increase. Thus, it is not guaranteed to early stop the training by using validation data.

\begin{figure}
\centering 
\includegraphics[width=.7\columnwidth]{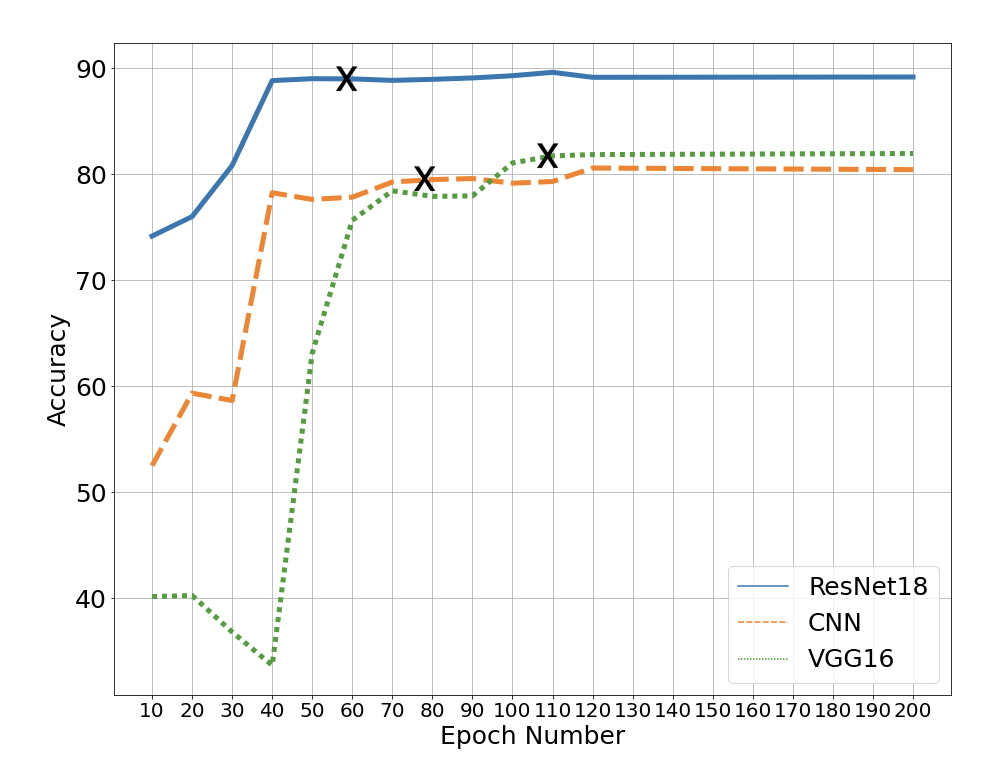}
\caption{The horizontal axis shows the epoch number (ranging from 10--200) used to train the ResNet18, CNN, and VGG16 on the CIFAR10 dataset. The vertical axis shows the testing accuracy of those models.  The X mark shows the testing accuracy and the epoch number to train a CNN variant based on the near-optimal learning capacity anticipated by our hypothesis (best viewed in color).} 
\label{fig:accuracy}
\end{figure}

We further analyze the generalization ability of CNN variants across a wide range of epoch numbers to train. Figure~\ref{fig:accuracy} shows the testing accuracy (top-1 classification) of ResNet18, VGG16, and CNN on the CIFAR10 dataset where 10 to 200 epochs are used for training.  Figure~\ref{fig:accuracy} shows that all the model's testing accuracy reaches a stable stage after a certain number of training epochs. Our hypothesis anticipates that ResNet18, VGG16, and CNN reach the near-optimal learning capacity at epochs 59, 109, and 78, respectively (marked by X). In summary, Figure~\ref{fig:accuracy} shows that the CNN variants generalization ability (i.e., the ability to predict on unseen data) does not significantly improve after the near-optimal learning capacity anticipated by our hypothesis. 

\section{Training behavior analysis}
We study the data variation across all the layers of CNN variants and datasets in the training phase. We examine the data variation after convolution operation by introducing the concept of stability vector ($S_n^e$) discussed in Section~\ref{sec:stabilityVector}. To anticipate the near-optimal learning capacity of CNN variants, we compare the mean of stability vector ($\delta_n^e$) with its previous epoch. In Table ~\ref{expTable}, we show the ablation study which supports our hypothesis. This section provides a detailed analysis of the pattern we observed in the training phase of CNN variants.

%To analyze our hypothesis, we study the data behavior at all the layers of CNN architectures during the training phase. We introduce data stability concept in layers of CNN which  identifies a CNN model's ability to learn from training data. Our hypothesis predicts the near optimal epoch number required to train a CNN model. We perform ablation study which supports our hypothesis.

\begin{figure}
\begin{tabular}{cc}
  \includegraphics[width=.48\linewidth]{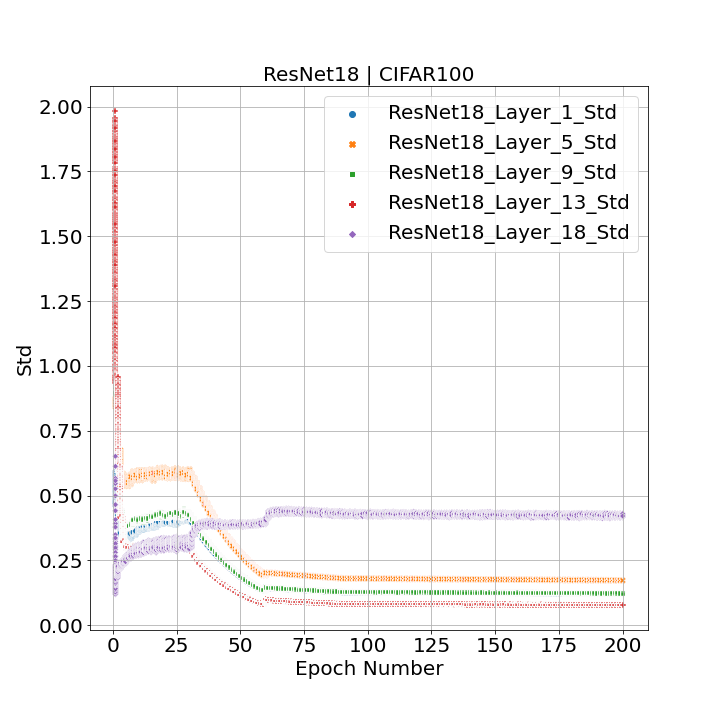}&
  \includegraphics[width=.48\linewidth]{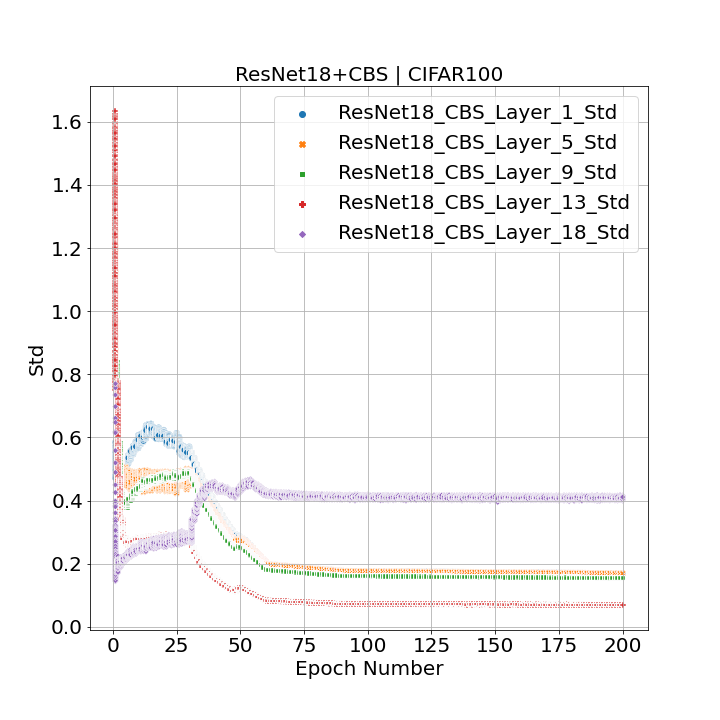} \\
  \includegraphics[width=.48\linewidth]{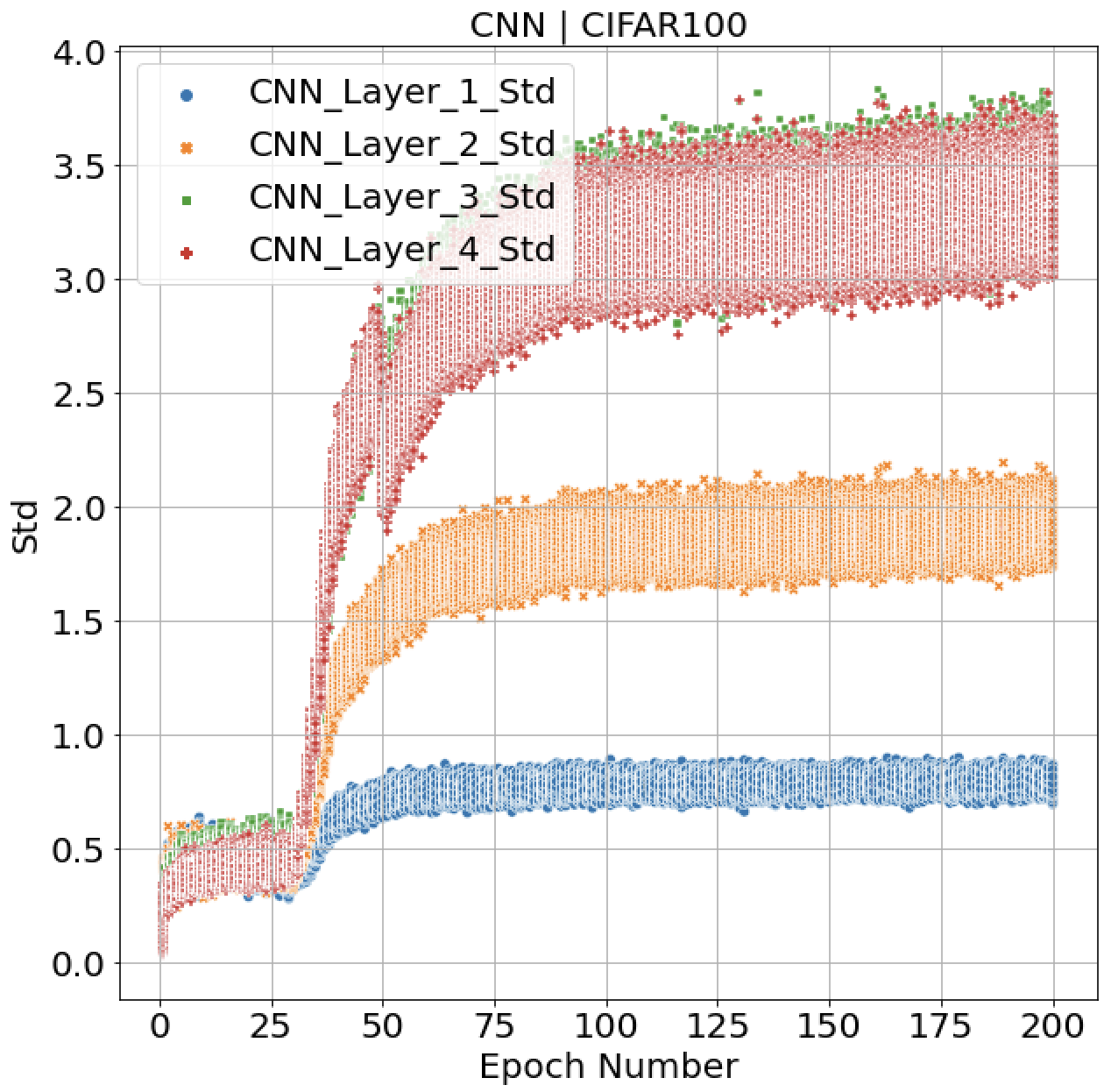} & 
  \includegraphics[width=.48\linewidth]{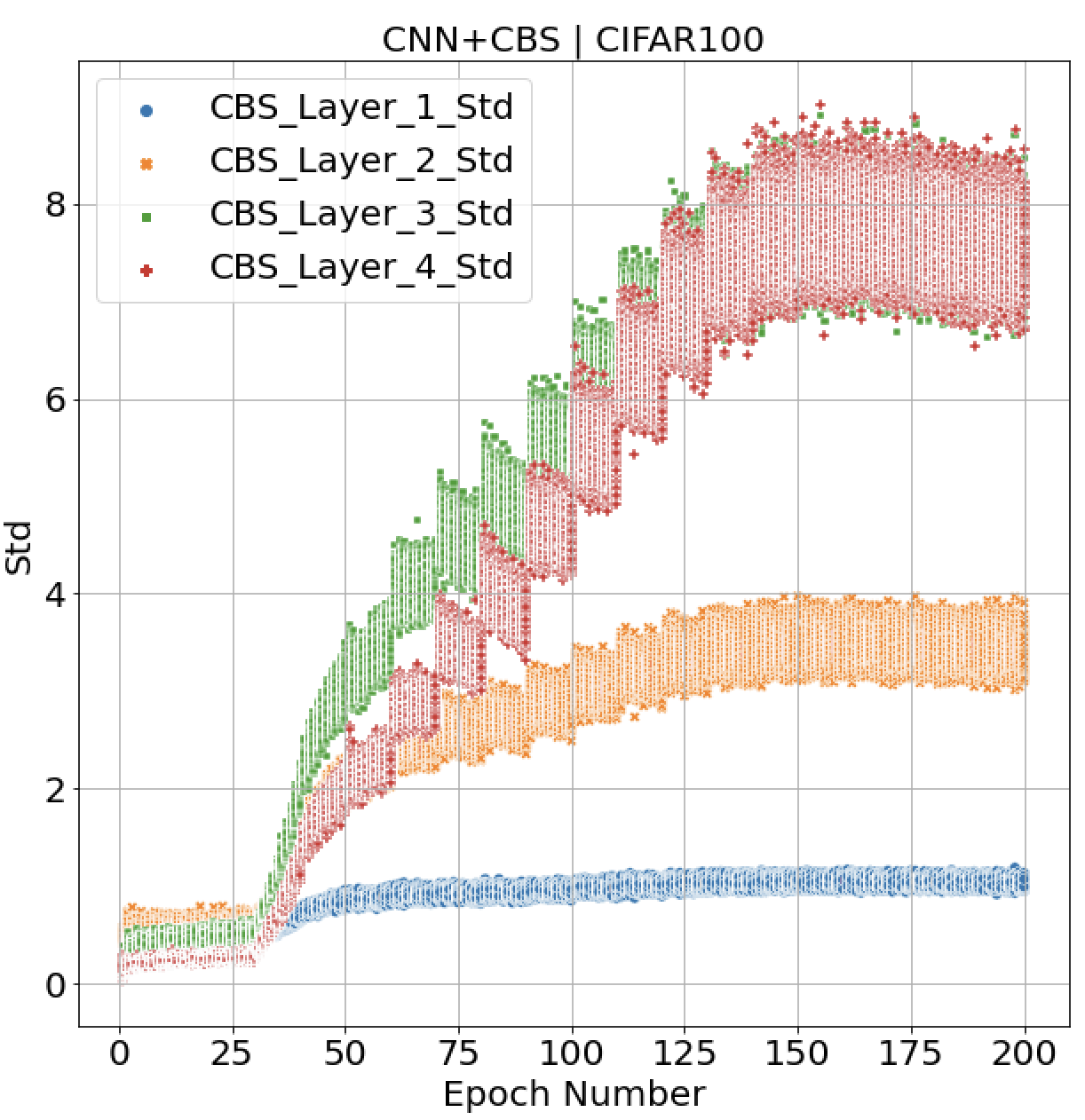} \\ 
  \includegraphics[width=.48\linewidth]{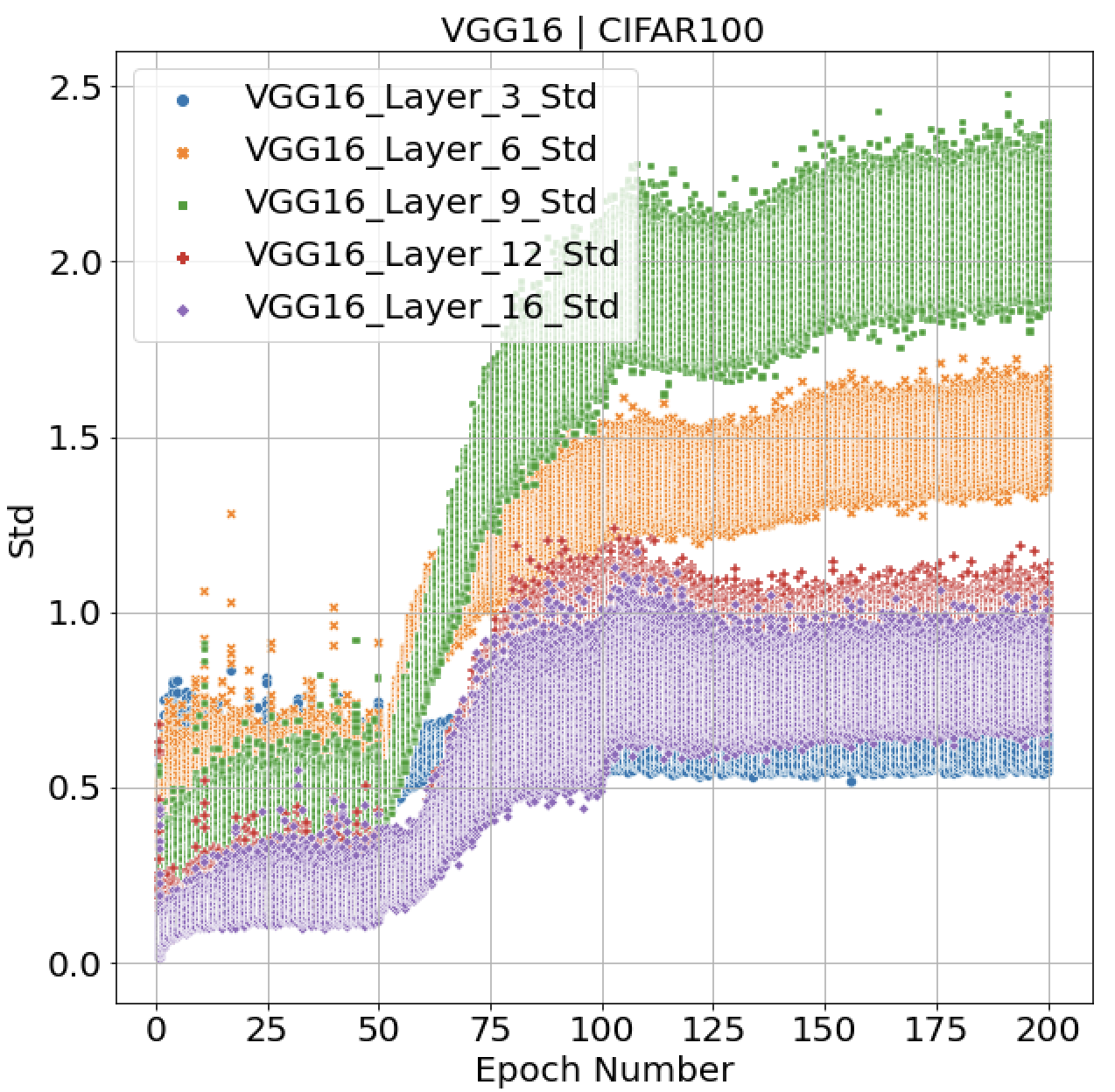} &   
  \includegraphics[width=.48\linewidth]{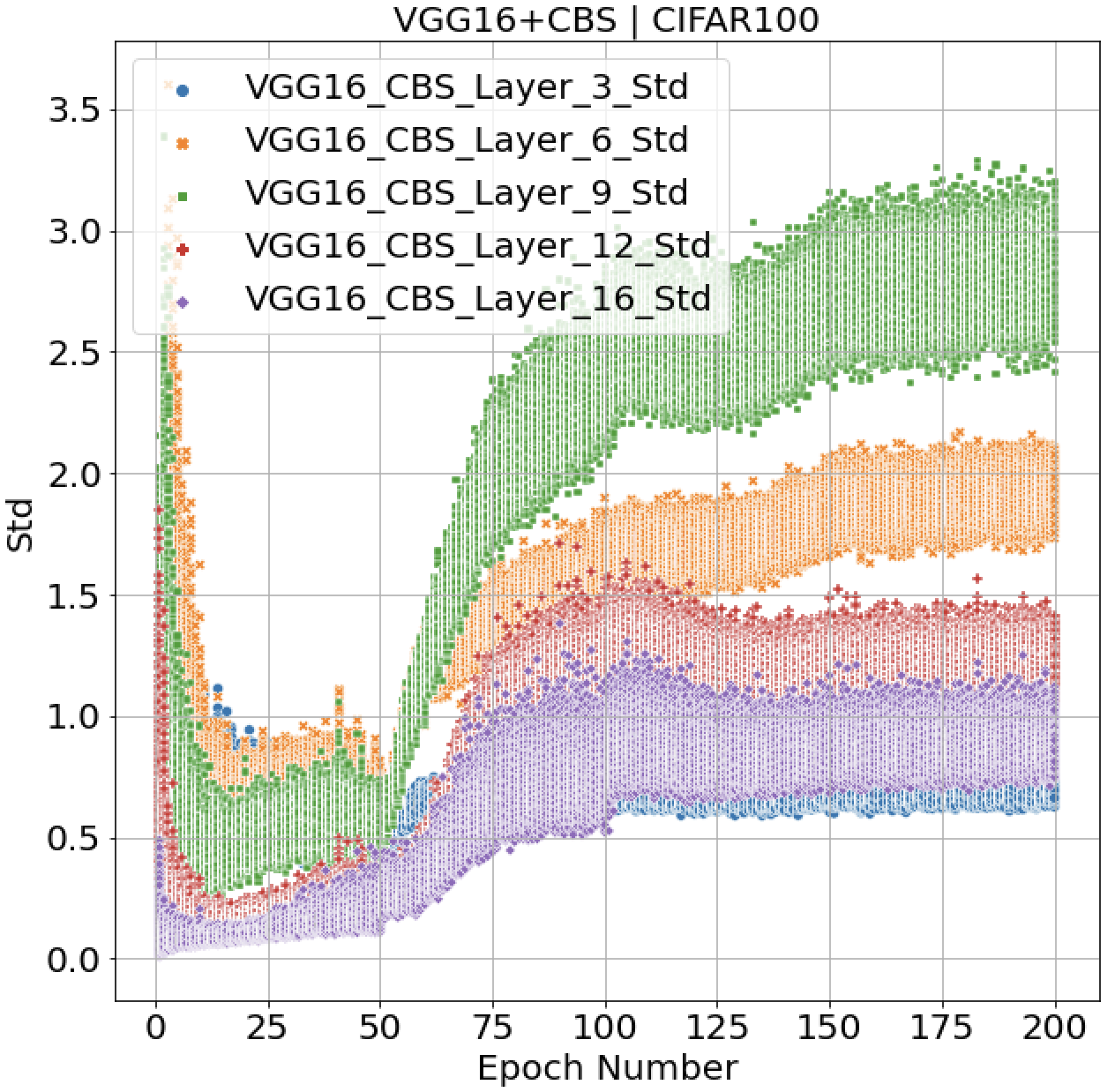} \\
\end{tabular} 
\caption{Data variation after convolution operation for different layers of ResNet18, CNN, VGG16 and their CBS variants on the CIFAR100 dataset for 200 epochs.}
\label{fig:cifar100}
\end{figure} 

\begin{figure*}[htp]
\begin{tabular}{cc}
  \includegraphics[width=.47\linewidth]{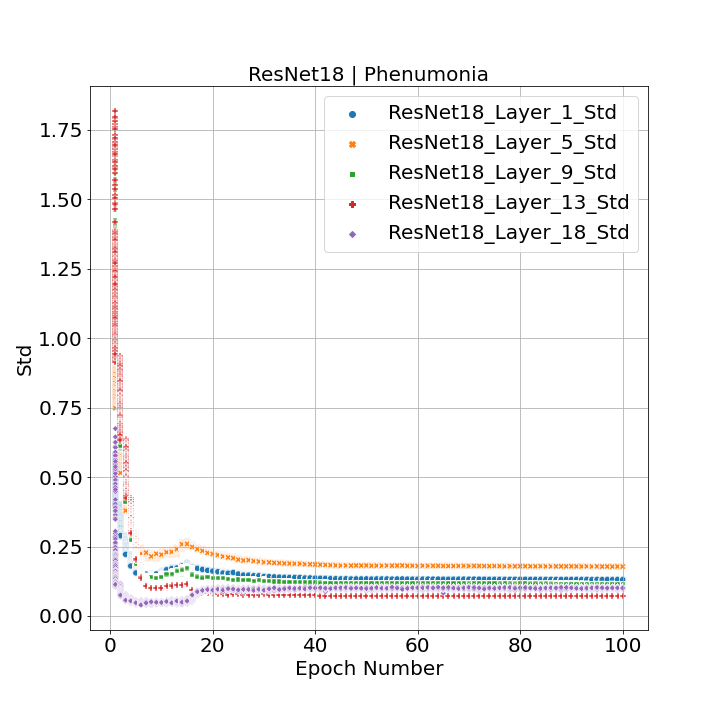}&   
  \includegraphics[width=.47\linewidth]{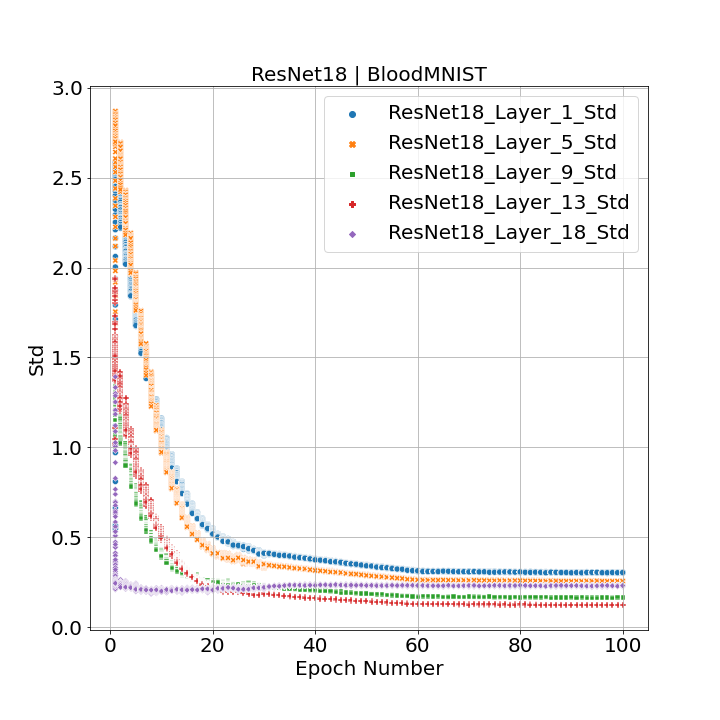} \vspace{-11pt}\\
  
  \includegraphics[width=.47\linewidth]{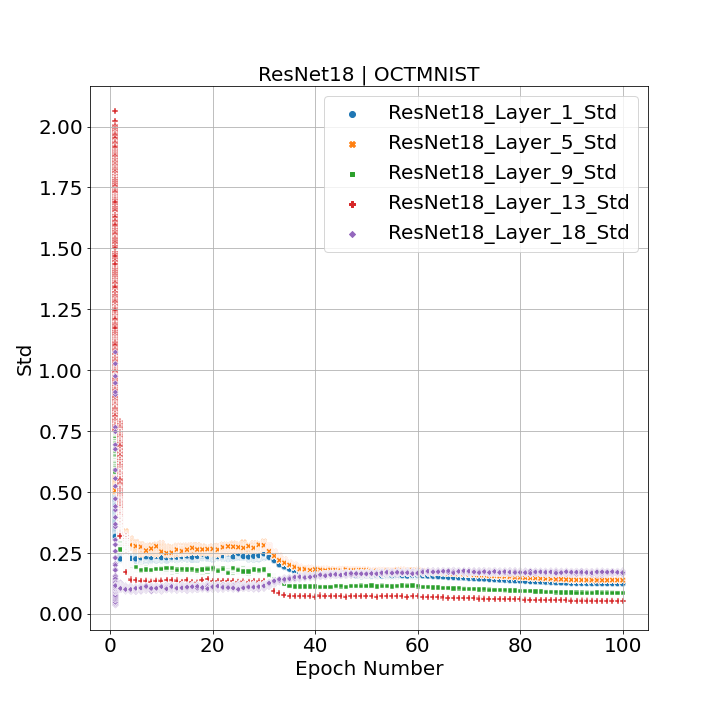}& 
  \includegraphics[width=.47\linewidth]{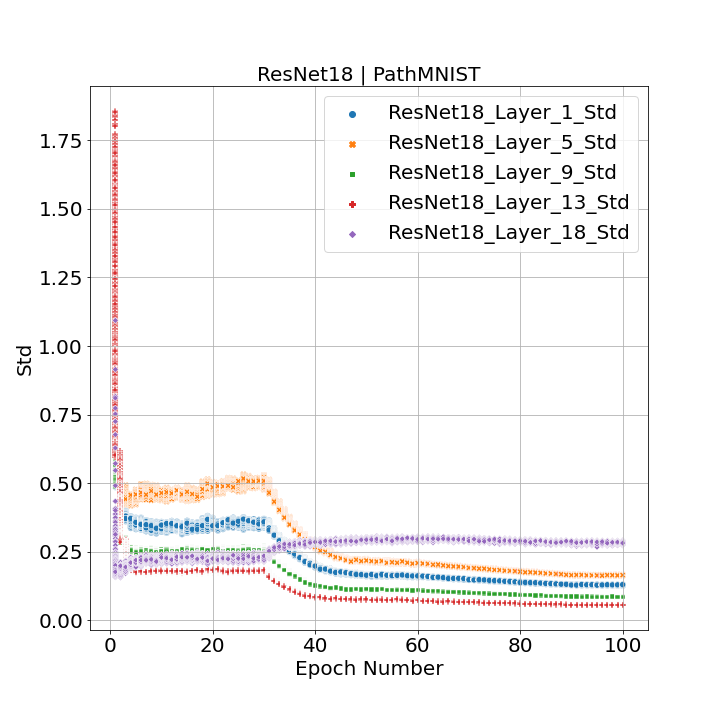}\\
  
  \includegraphics[width=.47\linewidth]{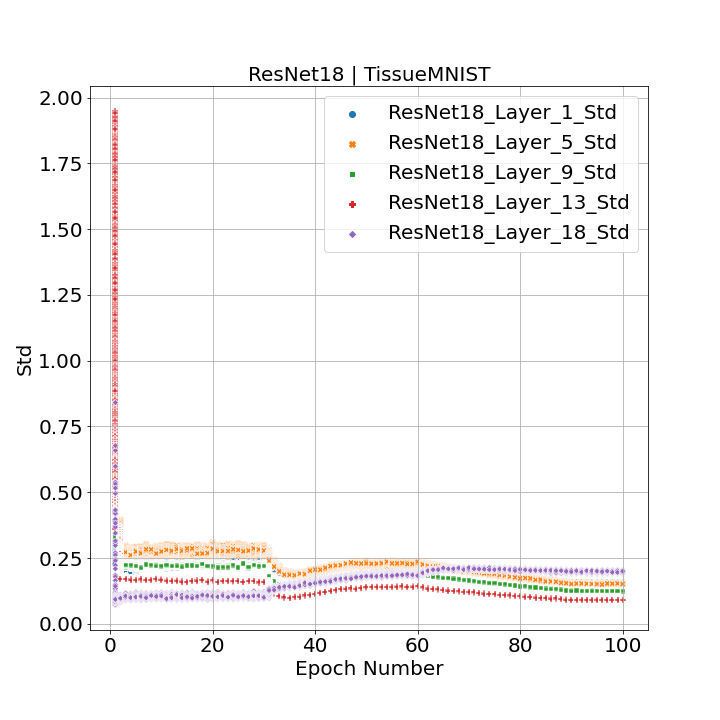}&
  \includegraphics[width=.47\linewidth]{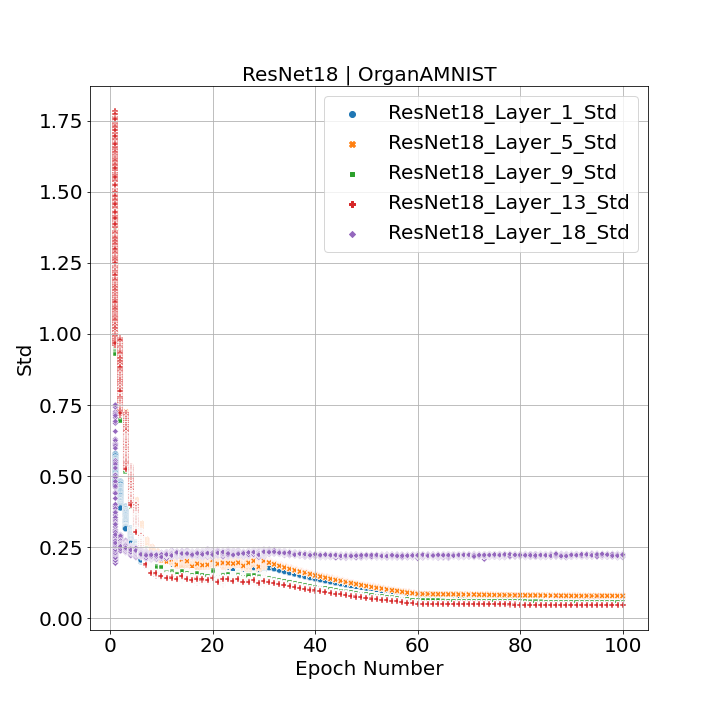}\\  
\end{tabular} \vspace{-11pt}
\caption{Data variation after convolution operation for number 1, 5, 9, 13 and 18 layer of ResNet18 on the Pneumonia detection, BloodMNIST, OCTMNIST, PathMNIST, TissueMNIST and OrganAMNIST dataset for 100 epochs.} \vspace{-10pt}
\label{fig:cifar100M}
\end{figure*}

Figure~\ref{fig:cifar100} shows the $S_n^e$ values for six different CNN variants on CIFAR100 datasets across 200 epochs. Figure~\ref{fig:cifar100M}  shows the $S_n^e$ values for ResNet18 on the Pneumonia detection, BloodMNIST, OCTMNIST, PathMNIST, TissueMNIST and OrganAMNIST dataset for 100 epochs.

As an example, top-left of Figure~\ref{fig:cifar100} shows the $S_n^e$ values (for layers 1, 5, 9, 13, and 18) for ResNet18 on CIFAR100 datasets across 200 epochs. Each epoch contains 782 data points (i.e., $t_{\text{train}}$ in Table~\ref{datasetTable}) for each layer. To explain the $S_n^e$ values behavior, we divide the training phase into the following four different phases to anticipate near-optimal learning capacity of CNN variants: Initial phase, curved phase, curved phase to stable phase, and stable phase. It is noteworthy that the range of all the phases can vary based on CNN variants and dataset as shown in Figure~\ref{fig:cifar100} and Figure~\ref{fig:cifar100M}.

\begin{figure}[htp]
\begin{subfigure}{1\textwidth}
  \centering 
  \makebox[\textwidth]{\includegraphics[width=1.2\textwidth ]{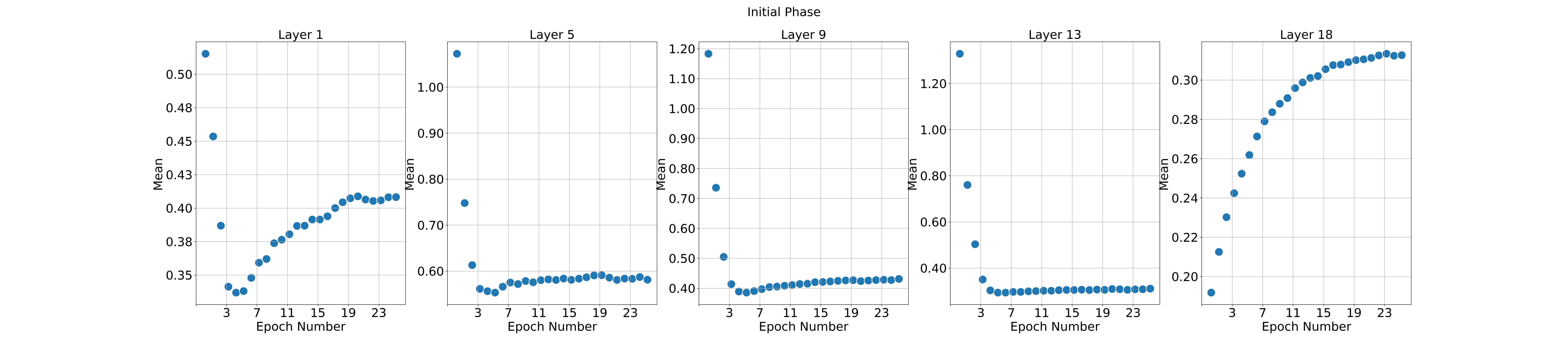}}
  \caption{$\mu_{n}^e$  values show the instability during the initial phase of training from epoch 1 to 25  for ResNet18 on CIFAR100 dataset. The instability shows for layer 1, 5, 9, 13, and 18. The sharp drop of $\mu_{n}^e$ values can be observed in the initial phase.}
  \label{fig:initial}
\end{subfigure}

\begin{subfigure}{1\textwidth}
  \centering 
  \makebox[\textwidth]{\includegraphics[width=1.2\textwidth ]{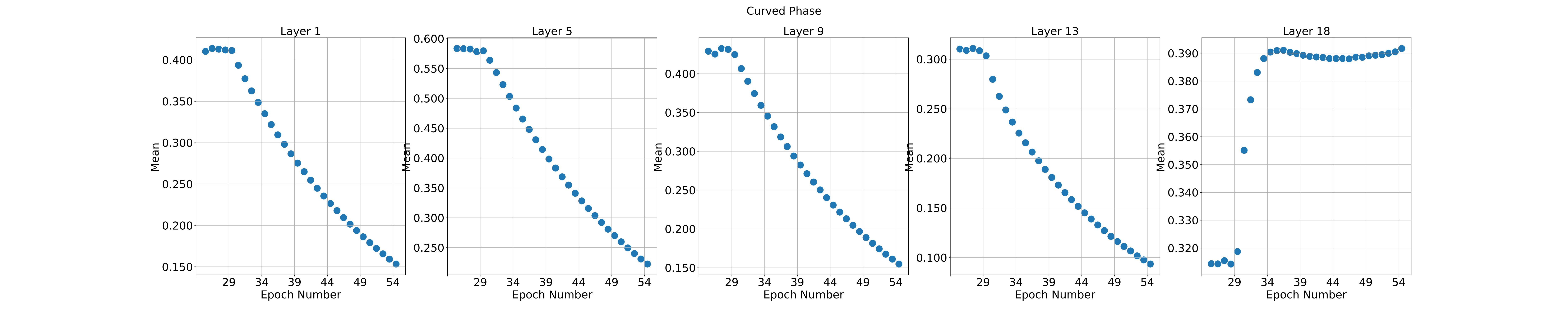}}
 \caption{$\mu_{n}^e$ values show the gradual increase or decrease from epoch 26 to 55 for ResNet18 on CIFAR100 dataset. This smooth transition of $\mu_{n}^e$ values creates a curved shape across all layers.}
  \label{fig:curve}
\end{subfigure}

\begin{subfigure}{1\textwidth}
  \centering
  \makebox[\textwidth]{\includegraphics[width=1.2\textwidth ]{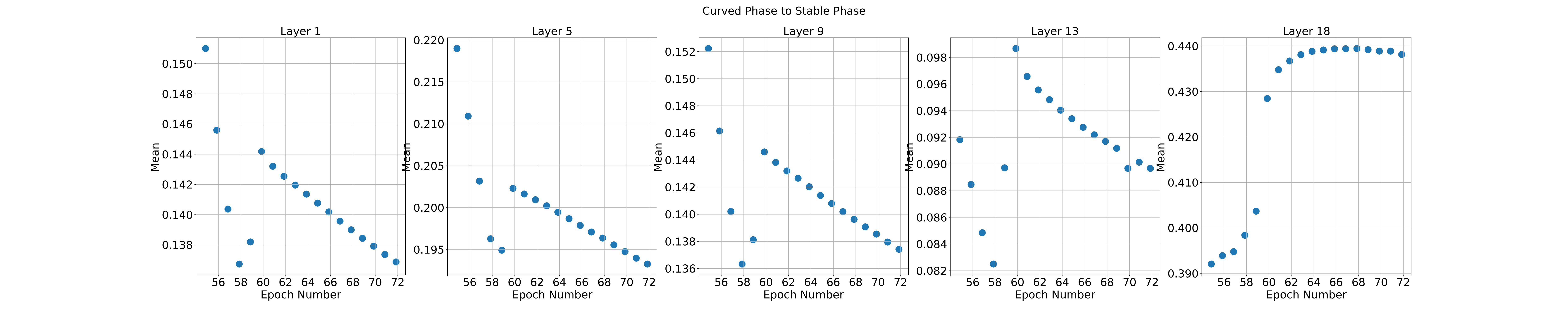}}
 \caption{$\mu_{n}^e$ values show significantly low fluctuation as the model gets closer to its near optimal learning capacity.}
  %$\mu_{n}^e$ values of training from epoch 56 to 72  for ResNet18 on CIFAR100 dataset.
  
  \label{fig:curveToStable}
\end{subfigure}

\begin{subfigure}{1\textwidth}
  \centering 
  \makebox[\textwidth]{\includegraphics[width=1.2\textwidth ]{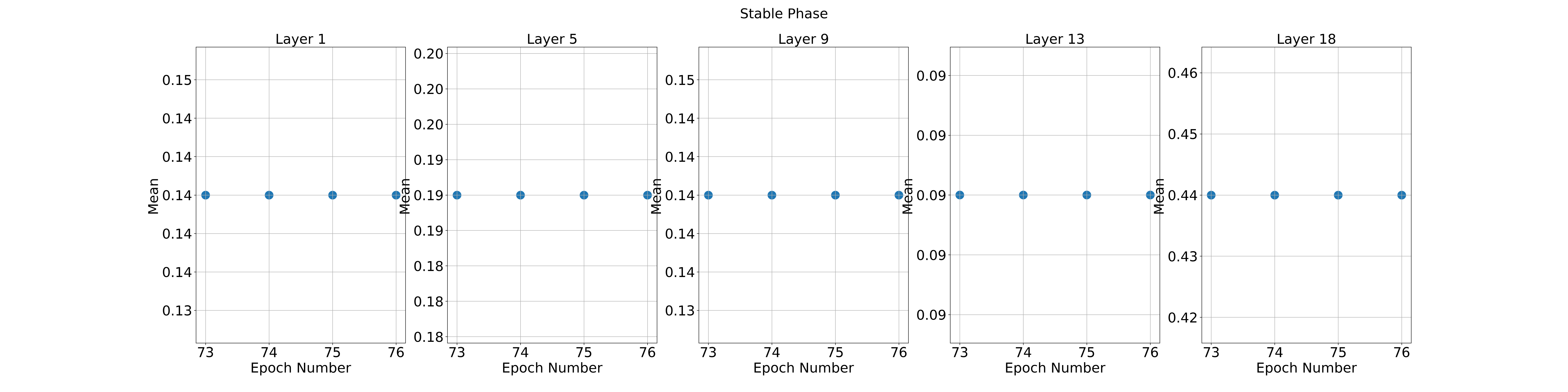}}
 \caption{As the rate of change of $\mu_{n}^e$ values gets significantly low, the probability of getting stable $p^2(\mu_{n}^e)$  values for consecutive epochs gets higher. At stable phase, $\delta_n^e$=0 indicates that the CNN reaches its near optimal learning capacity and terminates the training. Our hypothesis predicts 76 as the near optimal epoch number for ResNet18 on CIFAR100 dataset.}
  \label{fig:stable}
\end{subfigure}

\caption{Mean of stability values ($\mu_{n}^e$) at ResNet18 on the CIFAR100 dataset. Figures~\ref{fig:initial},~\ref{fig:curve}, and ~\ref{fig:curveToStable} show $\mu_{n}^e$ values on initial phase, curved phase, and curved phase to stable phase. Figure~\ref{fig:stable} shows $p^2(\mu_{n}^e)$ values on stable phase.}  
\label{fig:resnetAll}
\end{figure}

 At $e$-th epoch and $n$-th layer, each data point of Figure~\ref{fig:resnetAll} shows the $\mu_n^e$ value which measures the mean \st{of} stability vector ($S_n^e$). We also show the behavior of $\mu_n^e$ in those four phases. Our goal is to identify the `stable phase' to anticipate the near-optimal learning capacity of CNN variants. We use subplots for different layers in Figure~\ref{fig:resnetAll} to provide better understanding.

%Figure~\ref{fig:resnetAll} contains more detail explanation of our analysis during these four phases. Each data point represents the mean of stability vector ($\mu_n^e$). 

% \vspace{5pt}

 \textbf{Initial phase} refers to the early stage of training. For ResNet18 on the CIFAR100 dataset, we consider the approximate range of the initial phase from epoch 1 to epoch 25. In this phase, $S_n^e$ values are unstable across all the layers (top-left one of Figure~\ref{fig:cifar100}). We also observe a sharp drop or rise of $\mu_n^e$ values in most of the layers (Figure~\ref{fig:initial}).

%It is noteworthy that at layer 1, 5, 9, and 13, the $\mu_n^e$ values decrease while at 18-th layer the $\mu_n^e$ values increase with epochs (Figure~\ref{fig:initial},~\ref{fig:curve},~\ref{fig:curveToStable}). The reason behind this is that there is an average pool function used in the last layer of ResNet18.

\textbf{Curved phase} refers to the smooth changes of $S_n^e$ values in the training phase. For ResNet18 on the CIFAR100 dataset, we consider the curved phase's approximate range from epoch 26 to epoch 55. We observe $S_n^e$ values gradually increase or decrease (Figure~\ref{fig:cifar100}, top-left) in curved phase.  Figure~\ref{fig:curve} also shows that $\mu_{n}^e$ values across all the layers create a smooth-shaped curve.

 \textbf{Curved phase to stable phase} refers to the indication that CNN gets closer to its near-optimal learning capacity. For ResNet18 on the CIFAR100 dataset, we consider the curved phase to stable phase's approximate range from epoch 56 to epoch 72. At the start of this phase, the $\mu_{n}^e$ values fluctuate, but as the training goes on, the fluctuations gradually increase or decrease with epochs. Figure~\ref{fig:curveToStable} shows the $\mu_{n}^e$ values for ResNet18 on CIFAR100 dataset ranging epoch 56 to 72.

\textbf{Stable phase} refers to the range of epochs where the change of $\mu_n^e$ values are almost insignificant across all the layers. For each layer $n$, we compare the mean of stability vector with its previous epoch by rounding to decimal places $r$ using Equation~\ref{eq:precision} to compute $\delta_n^e$. If there is no significant difference between two epochs' mean of stability vectors for all the layers, in that case, it indicates the possibility that the CNN variant is close to its near-optimal learning capacity. To make sure that the CNN variant reaches its near-optimal learning capacity, we verify the $\sum_{i=1}^n \delta_i^e =0$ for two more epochs. Figure~\ref{fig:stable} shows the stable region for ResNet18 on the CIFAR100 dataset. In Figure~\ref{fig:stable}, we can observe that after two decimal points, there are no changes in $\mu_{n}^e$ values from epoch 73 to 76 for all $n$ layers. Thus, our hypothesis terminates the training of ResNet18 on the CIFAR100 dataset at epoch 76.

 It is noteworthy that in the stable phase, we compute $\delta_n^e$ by using the function $p^r$, and we choose the value of $r\!=\!2$. Choosing the value of $r\!=\!1$ causes a very early stop of the training, while $r\!=\!3$ does not guarantee stopping training at the near-optimal learning capacity. Choosing $r\!\!>\!\!3$ does not stop training even if the epoch number is large enough\footnote{We checked with $r\!=\!4$ for ResNet18 on the CIFAR100 dataset and VGG16 architecture on the SVHN dataset, and the models do not stop training even after the 350 epoch.}.
 
 We observe a similar pattern of data variation after convolution operation ($S_n^e$) for all the six CNN variants on CIFAR10, CIFAR100, and SVHN datasets. Figure~\ref{fig:cifar100} shows  $S_n^e$ values of these six CNN variants during the training phase on the CIFAR100 dataset.

\section{Conclusion}
In this paper, we analyze the data variation of a CNN variant by introducing the concept of stability vector to anticipate the near-optimal learning capacity of the variant. Current practices select arbitrary safe epoch numbers to run the experiments. Traditionally, for early stopping, validation error with train loss is used to identify the generalization gap. However, it is a trial-and-error-based approach, and recent studies suggest that train loss does not correlate well with generalization. We propose a hypothesis that anticipates the near-optimal learning capacity of a CNN variant during the training and thus saves computational time. The proposed hypothesis does not require a validation dataset and does not introduce any trainable parameter to the network. The implementation of the hypothesis can be easily integrated into any existing CNN variant as a plug-and-play. We also provide an ablation study that shows the effectiveness of our hypothesis by saving 58.49\% computation time (on average) across six CNN variants and three general image datasets. We also conduct our experiment by using our hypothesis on ten medical image dataset and save 44.1\% computational time compared to MedMNIST-V2 without losing accuracy. We expect to further investigate the data behavior based on different statistical properties for other deep neural networks.

\section{Conflict of interest}
louisiana.edu
\section{Data Availability Statement}
All the datasets used in the experiments are publicly available.

\backmatter

\bmhead{Supplementary information}

If your article has accompanying supplementary file/s please state so here. 

Authors reporting data from electrophoretic gels and blots should supply the full unprocessed scans for key as part of their Supplementary information. This may be requested by the editorial team/s if it is missing.

Please refer to Journal-level guidance for any specific requirements.

\bmhead{Acknowledgments}

Acknowledgments are not compulsory. Where included they should be brief. Grant or contribution numbers may be acknowledged.

Please refer to Journal-level guidance for any specific requirements.

\section*{Declarations}

Some journals require declarations to be submitted in a standardised format. Please check the Instructions for Authors of the journal to which you are submitting to see if you need to complete this section. If yes, your manuscript must contain the following sections under the heading `Declarations':

\begin{itemize}
\item Funding
\item Conflict of interest/Competing interests (check journal-specific guidelines for which heading to use)
\item Ethics approval 
\item Consent to participate
\item Consent for publication
\item Availability of data and materials
\item Code availability 
\item Authors' contributions
\end{itemize}

\noindent
If any of the sections are not relevant to your manuscript, please include the heading and write `Not applicable' for that section. 

\bigskip
\begin{flushleft}%
Editorial Policies for:

\bigskip\noindent
Springer journals and proceedings: \url{https://www.springer.com/gp/editorial-policies}

\bigskip\noindent
Nature Portfolio journals: \url{https://www.nature.com/nature-research/editorial-policies}

\bigskip\noindent
\textit{Scientific Reports}: \url{https://www.nature.com/srep/journal-policies/editorial-policies}

\bigskip\noindent
BMC journals: \url{https://www.biomedcentral.com/getpublished/editorial-policies}
\end{flushleft}

\begin{appendices}

\section{Section title of first appendix}\label{secA1}

An appendix contains supplementary information that is not an essential part of the text itself but which may be helpful in providing a more comprehensive understanding of the research problem or it is information that is too cumbersome to be included in the body of the paper.




\end{appendices}


\bibliography{sn-article}% common bib file

\end{document}